\RequirePackage{fix-cm}
\documentclass[twocolumn]{svjour3}          % twocolumn
\smartqed  % flush right qed marks, e.g. at end of proof
\usepackage{graphicx}
\usepackage{amssymb}
\usepackage{latexsym}
\usepackage{amsmath}
\usepackage{url}
\usepackage{xcolor}
\usepackage{multirow}
\usepackage{subcaption}
\usepackage{afterpage}
\usepackage{placeins}
\usepackage{natbib}

 \journalname{International Journal of Computer Vision}
\begin{document}

\title{ShadingNet: Image Intrinsics by\\ Fine-Grained Shading Decomposition}


\author{Anil S. Baslamisli\textsuperscript{1,2} \and
        Partha Das\textsuperscript{1} \thanks{\textsuperscript{1} denotes equal contribution,
        \textsuperscript{2} denotes corresponding author}                \and
        Hoang-An Le        \and
        Sezer Karaoglu \and
        Theo Gevers
}


\institute{A. S. Baslamisli \and P. Das \and H.A. Le \and S. Karaoglu \and T. Gevers \at
              University of Amsterdam,
              Science Park 904,
              Amsterdam 1098XH,
              The Netherlands\\
            %   \email{a.s.Baslamisli@uva.nl, p.das@uva.nl\textsuperscript{a}, hoang-an.le@uva.nl, s.karaoglu@3duniversum.com, th.gevers@uva.nl}           %  \\
           \and
            P. Das \and S. Karaoglu \and T. Gevers \at
            3D Universum, Science Park 400, Amsterdam 1098XH, The Netherlands\\
           \and
           A. S. Baslamisli \at
           \email{a.s.baslamisli@uva.nl}
}

% \institute{A. S. Baslamisli \and P. Das \and H.A. Le \and S. Karaoglu \and T. Gevers \at
%               University of Amsterdam, Science Park 904, Amsterdam 1098XH, The Netherlands
              
%             \and 
%             P. Das \and S. Karaoglu \and T. Gevers \at
%             3D Universum, Science Park 400, Amsterdam 1098XH, The Netherlands
%             %   \email{fauthor@example.com}           %  \\
% %             \emph{Present address:} of F. Author  %  if needed
%         %   \and
%         %   P. Das \at
%         %       \email{p.das@uva.nl}
% }

\date{Received: date / Accepted: date}

\maketitle

\begin{abstract}
In general, intrinsic image decomposition algorithms interpret shading as one unified component including all photometric effects. As shading transitions are generally smoother than reflectance (albedo) changes, these methods may fail in distinguishing strong photometric effects from reflectance variations. Therefore, in this paper, we propose to decompose the shading component into direct (illumination) and indirect shading (ambient light and shadows) subcomponents. The aim is to distinguish strong photometric effects from reflectance variations. An end-to-end deep convolutional neural network (ShadingNet) is proposed that operates in a fine-to-coarse manner with a specialized fusion and refinement unit exploiting the fine-grained shading model. It is designed to learn specific reflectance cues separated from specific photometric effects to analyze the disentanglement capability. A large-scale dataset of scene-level synthetic images of outdoor natural environments is provided with fine-grained intrinsic image ground-truths. Large scale experiments show that our approach using fine-grained shading decompositions outperforms state-of-the-art algorithms utilizing unified shading on NED, MPI Sintel, GTA V, IIW, MIT Intrinsic Images, 3DRMS and SRD datasets.
\end{abstract}

\section{Introduction}
\label{sec:intro}
Intrinsic image decomposition aims to recover the image formation components in terms of reflectance (albedo) and shading (illumination) \citep{Barrow1978}. The reflectance component contains information about the real color (i.e. albedo) of an object and is independent of illumination and camera viewpoint. The shading component contains different types of photometric effects such as direct light, ambient light (inter- and intra- reflections) and shadow casts. As a result, using intrinsic images rather than raw $RGB$ images can be favourable for different computer vision tasks. For instance, reflectance images (i.e. illumination invariant) are useful for semantic segmentation task for scene understanding \citep{Baslamisli2018ECCV}. They are also preferred by computational photography applications for plausible photo editing tasks such as recoloring, material editing and retexturing \citep{Meka2016}. Even recently, the textile industry favors them for improved fabric recolorization \citep{Xu2019}. On the other hand, shading images are a source of information for 3D shape reconstruction tasks \citep{Wada1995,Henderson2020}, and for color constancy \citep{Gijsenij2008}.

The problem of intrinsic image decomposition is ill-posed, because there can be multiple solutions to reflectance and shading that reconstruct the same input. As a consequence, most of the traditional methods impose priors on the intrinsic components to constrain the search space by means of an optimization process \citep{Gehler2011,Shen2011,Barron2015}. Recent approaches use large scale datasets with powerful deep learning methods \citep{Shi2017,Baslamisli2018CVPR,Li2018ECCV}. 

In general, most of the intrinsic image decomposition methods (traditional and new ones) assume a single unified shading component containing all the photometric effects. The common assumption is that strong image variations are due to reflectance changes and that smooth image variations are caused by shading. However, this assumption does not always hold for real images as they may suffer from strong photometric changes due to environmental conditions such as heavy shadow casts and inter-reflections. In fact, it is demonstrated that intrinsic image decomposition methods perform poorly in handling shadow casts \citep{Isaza2012}. Altering strong shadow casts by reflectance variations may negatively influence the quality of the resulting intrinsic image decomposition.

Therefore, our goal is to represent the different photometric effects separately into direct (i.e. light source) and indirect light (i.e. ambient light and shadow cast) components. The aim is to explicitly model photometric effects that may cause drastic changes in pixel values to provide extra cues to the reflectance map estimations for better disentanglement of color changes from those strong intensity variations. To this end, we extend the standard image formation model to decompose the shading component into direct (light source) and indirect light conditions (ambient light and shadow casts). Based on the fine-grained model, an end-to-end deep convolutional neural network (CNN) is proposed that operates in a fine-to-coarse manner with a specialized fusion and refinement unit. More precisely, the contributions of our work are as follows:
\begin{itemize}
    \item We propose ShadingNet, the first end-to-end model for learning the fine-grained shading decompositions (photometric effects) of natural scenes (Figure~\ref{fig:shadingNet_model}).
    \item We specifically design the model to couple a shading decomposition with a reflectance prediction to learn specific reflectance cues separated from specific photometric effects to analyze the disentanglement capability.
    \item We propose a generic rendering pipeline to generate fine-grained shading decompositions. We demonstrate it using Blender Cycles and extend a subset of the Natural Environments Dataset (NED) \citep{Sattler2017,Baslamisli2018ECCV,Le2020}\footnote{The models and the dataset will be made publicly available.}.
    \item We systematically analyze the quality and contributions of the fine-grained shading decompositions using quantitative and qualitative evaluations on seven different datasets (NED, MPI Sintel, GTA V, IIW, MIT Intrinsic Images, 3DRMS and SRD), achieving superior performance compared with state-of-the-art models estimating a unified shading map. 
\end{itemize}

\section{Related Work}
\label{sec:related_work}
Intrinsic image decomposition is an ill-posed and under-constrained problem, because different reflectance and shading maps can reconstruct the same input. Traditional work usually aims to constrain the search space by imposing priors on the intrinsic components. One of the pioneering work is the Retinex algorithm \citep{Land1971}. It assumes that reflectance changes cause large gradients, whereas shading variations result in smaller ones. Since then, many priors have been introduced to approach the problem, such as reflectance sparsity \citep{Gehler2011,Shen2015}, texture \citep{Shen2008,Zhao2012}, depth \citep{Lee2012,Barron2013} and infrared \citep{Cheng2019}. It is also shown that using image sequences (video) is favorable for intrinsic image decomposition as it imposes a constant reflectance prior and varying shading for the same pixels within the sequence \citep{Weiss2001}. On the other hand, recent research generally use large-scale datasets and supervised CNNs. DirectIntrinsics is the first work that directly regresses reflectance and shading maps from $RGB$ images \citep{Narihia2015}. Since then, many deep learning based methods are proposed. For instance, ShapeNet model exploits the correlations between the intrinsic components by interconnecting decoder features \citep{Shi2017}. \citet{Baslamisli2018CVPR} consider both a physics-based reflection model and intrinsic gradient supervision to steer the learning process. \citet{Lettry2018b} utilize adversarial learning, and CGIntrinsics leverage 4 different datasets for better intrinsic image decompositions \citep{Li2018ECCV}. \citet{Baslamisli2020} steers the learning process by physics-based invariant descriptors. \citet{Fan2018} incorporates a guidance network and a domain filter to obtain realistically flattened reflectance images. The task is also approached in an unsupervised fashion using a single $RGB$ image \citep{Liu2020,Liu2020b}. In addition, image sequences over time are exploited to constrain the reflectance also within deep learning frameworks \citep{Lettry2018a,Li2018CVPR}. Finally, recent works on inverse scene rendering also aim at estimating scene-level reflectance maps \citep{Sengupta2019,Yu2019,Li2020}.

Most of the intrinsic image decomposition algorithms represent shading as one unified component including all photometric effects. Nonetheless, there are a number of optimization-based methods that disentangle the shading problem by performing additional decompositions. For instance, the illumination image can be separated into direct and multiple indirect components for plausible material coloring \citep{Carroll2011}. However, this method requires additional user strokes. SIRFS recovers shape, reflectance and chromatic illumination \citep{Barron2015}, but it performs poorly on natural scenes. An improved version requires depth information \citep{Barron2013}. \citet{Laffont2013} propose a model that not only separates reflectance from illumination, but also factorizes the illumination into sun, sky and indirect layers. The model requires multiple views of the same scene. In another work, an image is decomposed into reflectance, shading, direct irradiance, indirect irradiance and illumination color by using a multiplicative model that prevents further decomposition of shadow casts and ambient light \citep{Chen2013}. Likewise, their method is dependent on depth information. On the other hand, there are a few deep learning based methods that perform additional decompositions. \citet{Janner2017} decomposes single images into reflectance, shape, and lighting maps. However, instead of modelling the photometric effects, they approximate the shading process of a rendering engine, which again aims at estimating a unified shading component. \citet{Innamorati2017} decompose an object centered image to reflectance, occlusion, diffuse illumination, specular shading and surface normals for user friendly photo editing. Finally, GLoSH predicts global spherical harmonics for lighting, reflectance and surface normals \citep{Zhou2019}. In contrast to existing methods, we propose to decompose (scene-level) shading into direct and indirect shading terms to model the different photometric effects without any specialized regularization or additional sensory information such as depth. We further factorize the indirect shading term to model ambient light and shadow casts, whereas the direct shading is defined by object geometry and light source interactions. The aim is to explicitly model photometric effects to provide specific cues to the reflectance map estimation for better disentanglement of color changes from those strong photometric intensity variations.

Recently, supervised-based CNN methods use large-scale datasets \citep{Shi2017,Baslamisli2018ECCV,Li2018ECCV,Sengupta2019}. Outdoor scenes are frequently influenced by strong shadow casts and varying lighting conditions. Unfortunately, existing datasets lack variations of these types of photometric effects, except for the Natural Environments Dataset (NED) \citep{Sattler2017,Baslamisli2018ECCV,Le2020}. Other datasets are either object centered or taken from indoor scenes. NED contains natural (outdoor) environments under varying illumination conditions with dense intrinsic image ground-truths. We extend a subset of this dataset to generate direct shading (shading due to surface geometry and illumination conditions), shadow casts, and ambient light (inter-reflections) ground-truth images ($\approx30k\;images$). Additionally, we provide a generic rendering pipeline to generate those fine-grained shading decompositions.

\section{Fine-Grained Shading Decomposition}
\label{sec:extended_decomposition}

\subsection{Standard Image Formation Model}
\label{sec:ifm}
We use the Lambertian component of the dichromatic reflection model as the basis of our image formation \citep{Shafer1985}. Then, an image \emph{I} over the visible spectrum $\omega$ is modelled by:

\begin{equation} \label{eq:imf}
I = m(\vec{n}, \vec{l}) \int_{\omega}^{} \rho_{b}(\lambda)\; e(\lambda)\;f(\lambda)\; \mathrm{d}\lambda \;,
\end{equation}

\noindent where $\vec{n}$ indicates the surface normal, $\vec{l}$ denotes the (direct) light source direction, and \emph{m} is a function of the geometric dependencies (e.g. Lambertian $\vec{n} \cdot \vec{l}$). Furthermore, $\lambda$ represents the wavelength, $f$ indicates the camera spectral sensitivity, and $e$ describes the spectral power distribution of the illuminant. Finally, $\rho_b$ denotes the reflectance i.e. the albedo (intrinsic color). Then, assuming a linear sensor response, a single light source, and narrow band filters, the equation can be simplified as follows:

\begin{equation} \label{eq:iid}
I = \rho  s_u \;,
\end{equation}

\noindent where an image \emph{I} can be modelled by a product of its unified shading $s_u$ and reflectance $\rho$ components. If the light source $e$ is colored, then that color information is embedded in the illumination (shading) component.  In general, in the context of intrinsic image decomposition, the shading component $s_u$ is only defined for \emph{direct} light (i.e. no occlusion) as follows:

\begin{equation}\label{eq:shading}
    s_d = e_d \; (\vec{n} \cdot \vec{l})\;,
\end{equation}

\noindent where $e_{d}$ is the intensity of the light source. Obviously, Equation~\eqref{eq:shading} does not include photometric effects such as ambient light or shadow casts. However, this assumption is often violated for real images. To compute intrinsic images, explicitly modelling these photometric effects may help correctly distinguish strong shadow cues causing drastic changes in pixel values from reflectance variations.

\subsection{Image Formation Model with Composite Shading} 
\label{sec:fine_ifm}
To incorporate the photometric effects of the ambient lights, and assuming that the ambient light is uniform, we use a linear function to model the relationship between direct and indirect light:
\begin{equation}\label{eq:FullModel}
    I = \rho e_d \; (\vec{n} \cdot \vec{l}) + \rho e_a \;,
\end{equation}
where $e_a$ is the intensity of ambient. To model a cast shadow, an occluder is used that blocks the direct illumination $e_d$ entirely and a portion $\alpha_S$ of the ambient light. The shadowed intensity is:
\begin{equation}
I^{S} = \alpha_S \rho e_a = \rho e_a (1 - \alpha_S) = \rho (e_a - \alpha_S e_a) \;,
\end{equation}
where, for an occluder, $\alpha_S$ indicates the fraction of reduced ambient light caused by the cast shadow. We denote $e_a$ by $e^{+}_a$ and $- \alpha_S e_a$ by $e^{-}_a$, then we arrive at a linear function to model the relationship between direct light, ambient light and shadow casts by:
\begin{equation}\label{eq:FullModel}
    I = \rho e_d \; (\vec{n} \cdot \vec{l}) + \rho e^{+}_a  + \rho e^{-}_a \;.
\end{equation}

The indirect light $e_i = e^{+}_a + e^{-}_a$ consists of ambient light, denoted by $e^{+}_a$, resulting in an additive term. Shadows are modelled by a negative term $e^{-}_a$. Ambient light $e^{+}_a$ causes objects to appear brighter, whereas shadows $e^{-}_a$ cause objects to appear dimmer. 
%Finally, we obtain the composite shading model:
%\begin{equation}\label{eq:ShadingModel}
%     I = \rho s\;,
%\end{equation}
%where the fine-grained shading component $s = e_d (\vec{n} \cdot %\vec{l}) + e^{+}_a + e^{-}_a$ distinguishes the Lambertian (direct) %shading and two (indirect) photometric effects.

\begin{figure}[t]
\includegraphics[width=1\linewidth]{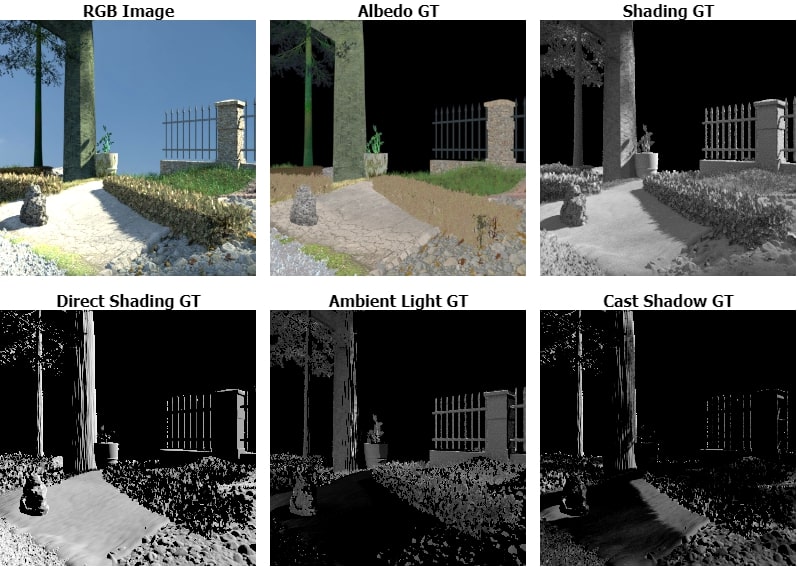}
\centering
\caption{A sample scene from NED with ground-truth intrinsics and fine-grained shading components. Ambient light further illuminates parts that direct light cannot reach. Shadow casts occur when direct light is occluded.}
\label{fig:dataset}
\end{figure}

\section{Dataset}
\label{sec:dataset}
\subsection{Natural Environments Dataset (NED)}
\label{sec:ned}
To train our models and baselines, we extend a subset of the (synthetic) Natural Environment Dataset (NED) introduced by \citet{Sattler2017,Baslamisli2018ECCV} to generate reflectance, direct shading (shading due to surface geometry and illumination conditions), ambient light and shadow cast ground-truth images. The dataset contains garden/park like natural (outdoor) scenes including trees, plants, bushes, fences, etc. Furthermore, scenes are rendered with different types of terrains, landscapes, and lighting conditions. Additionally, real HDR sky images with a parallel light source are used to provide realistic ambient light. Moreover, light source properties are designed to model daytime lighting conditions to enrich the photometric effects. Figure~\ref{fig:dataset} illustrates a sample scene from the dataset with dense ground-truth annotations. For the experiments, the dataset is randomly (scene) split resulting 15 gardens for training, around 25k images, and 3 gardens for testing, around 5k images.  

\subsection{Fine-grained Shading Rendering Pipeline}
\label{sec:renderer}
We re-render the aforementioned 18 gardens to obtain fine-grained shading components with dense ground-truth annotations. Scenes are re-rendered using physics-based Blender Cycles engine\footnote{\url{https://www.blender.org/}}. The rendering pipeline is modified to output reflectance and (unified) shading ground-truth intrinsic images, ground-truth surface normal images, and light source properties (color, position, and intensity). Then, we use Lambert's law to form the direct shading ($s_{d}$) component by Equation~\eqref{eq:shading}. Since Blender Cycles engine is modified to output surface normals and light source properties, ground-truth direct shading component is computed using Equation~\eqref{eq:shading}.

Then, the ground-truth indirect light effects (i.e. ambient light and shadow casts) are created. For the task, we use the ground-truth unified shading component, which is already made available by the rendering engine. Ambient light is due to extra light present on top of the direct shading, while shadow casts cause reductions in intensity values. As a result, subtracting the direct shading ground-truth from the unified shading ground-truth, we are left with indirect light effects that are modelled on top of the direct shading component. After subtraction, the resulting component has both positive (due to extra indirect light) and negative (due to lack of direct light) pixel values. We classify positive values as ambient light, whereas negative values are classified as shadow casts. Therefore, the procedure labels a pixel based on the dominant indirect light cue. Note that a pixel is not classified as either \emph{in} shadow or not, but has a continuous value. In that sense, umbra and penumbra regions can also be observed in the shadow maps depending on the intensity. Nonetheless, the formulation can be modified to facilitate a binary shadow map using a threshold. Finally, the input image is constructed by element-wise multiplying the unified shading and reflectance components to obtain a composite image that follows the physics-based image formation model.  

\section{Method}
\label{sec:method}
\subsection{ShadingNet}
\label{sec:shadingNet}
\begin{figure}[t]
\includegraphics[width=1\linewidth]{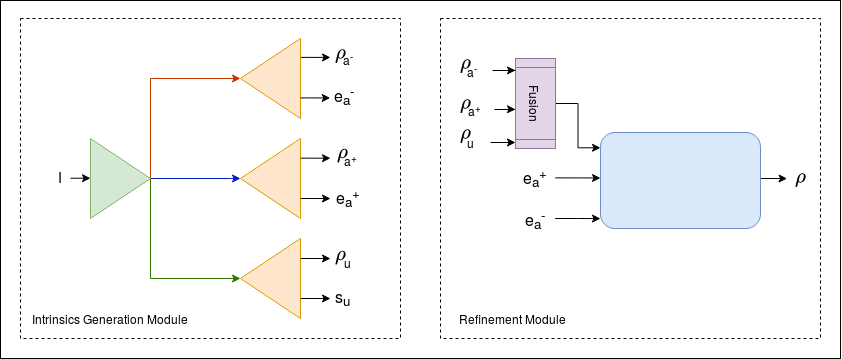}
\centering
\caption{ShadingNet model architecture. Each decoder tightly couples a reflectance prediction with a shading intrinsic in the \emph{Generation Module}. Learnable soft attention modules are applied to the encoder bottleneck features before forwarding to the relevant decoders to enhance feature discriminability. The \emph{Fusion Module} combines reflectance predictions with $1\times1$ convolutions as learnable weighted averages. The fusion is combined with the photometric cues and is fed to the \emph{Refinement Module} to generate the final reflectance map.}
\label{fig:shadingNet_model}
\end{figure}

Using the image formation model of Equation~\eqref{eq:FullModel}, we propose \emph{ShadingNet} to learn the fine-grained shading components for natural outdoor scenes. The model not only estimates the photometric effects, but also refines the reflectance predictions with a specialized fusion and refinement unit in a coarse-to-fine manner. It is illustrated in Figure~\ref{fig:shadingNet_model}. The generation module has one encoder and three decoders. The decoders generate unified shading ($s_u$), shadow cast ($e^{-}_a$), ambient light ($e^{+}_a$) and related reflectance predictions ($\rho_{u}$, $\rho_{a^{-}}$, $\rho_{a^{+}}$). To enhance feature discriminability and forward most relevant features to the decoders, soft attention modules are adapted on the encoder bottleneck. Thus, each decoder receives specialized bottleneck features.

Each decoder tightly couples a reflectance prediction with a shading intrinsic to further enforce feature discriminability. For example, the shadow decoder is designed to disentangle shadow \emph{and} reflectance cues. Thus, the aim of each decoder is to learn specific reflectance cues separated from specific photometric effects and from the unified shading. This rationale allows us to analyze the disentanglement quality of each photometric effect over the reflectance predictions.

Finally, the estimated fine-grained photometric components and the reflectance predictions are fused and refined to generate the final reflectance map. Since the reflectance images are predicted with different conditions and parameters in the decoders, they are expected to be not identical. That is exploited by the fusion unit aiming at generating learnable weighted combinations of the different reflectance images. Then, in contrast to the generation module which provides feature level cues and conditions the reflectances on the photometric cues as outputs, the refinement module conditions the fused reflectances on the photometric cues as inputs for full exploitation of the photometric effects.

Since our main motivation is to explicitly distinguish strong photometric effects from reflectance variations, we do not predict the direct shading component. Nonetheless, it is used as an indirect supervision signal, which is explained in the next section. The entire model is end-to-end trainable.

\subsubsection{Network Details}
\label{sec:implementation}
\textbf{Intrinsics Generation Module.} The encoder block uses 2-strided convolution layers for downsampling (5 times), except for the initial convolution layer. Each convolution is followed by 4 consecutive residual blocks \citep{He2016}. A residual block is composed of Batch Norm-ReLu-Conv($3\times3$) sequence, repeated twice. The bottleneck of the encoder is fed to 3 distinct efficient channel attention modules \citep{Wang2020}, which are then individually fed to the decoders. The decoders use Conv($3\times3$)-Batch Norm-LeakyReLu sequence. The feature maps are bilinearly up-sampled and concatenated with their encoder counterparts by skip connections \citep{Mao2016}. The process is repeated 5 times to reach the final resolution.

The initial encoder block takes a single input image of 3 channels ($RGB$) and produces 16 feature maps. Then, the number of feature maps is doubled for each following convolution operation until the penultimate block. Thus, the bottleneck has 256 feature maps. Efficient channel attention blocks use a kernel size of 5 and generate 256 feature maps. Then, each decoder receives specialized bottleneck features. Decoders generate 128 feature maps each, except the last layer that generates a single channel or three channels depending on the intrinsic component. All kernel weights are initialized using He initialization \citep{He2015}. The slope for the LeakyReLus is set to 0.01. The final outputs are passed through ReLus to avoid negative pixel values.
\newline \newline
\noindent\textbf{Fusion.} The three reflectance predictions are combined with learnable $1\times1$ convolutions to generate weighted combinations. The kernel weights are initialized using He initialization. The module takes 3 reflectance images as inputs (9 channels) and generates 24 channels combination. They are then concatenated with the photometric effects and fed to the final refinement network. 
\newline \newline
\noindent\textbf{Refinement Module.} The module takes input as the fused intrinsics and the photometric effects and outputs a single final reflectance map. Unlike the generation module which provides feature level cues, the refinement module conditions the fused reflectances on the photometric cues as inputs. The module starts with a convolution layer followed by 6 consecutive residual blocks with dilations $(2-2-4-8-8-1)$. Similar to the generation module, a dilated residual block is composed of Batch Norm-ReLu-Conv($3\times3$) sequence, repeated twice. The kernel weights are initialized using He initialization. The module does not involve any downsampling or upsampling operations. Each layer generates 32 features maps, except the last layer that predicts 3 channel reflectance image. The final output is passed through a ReLu to avoid negative pixel values. Since the entire model is trained end-to-end, the refinement module also improves the estimations of the generation module by back-propagation. 

\subsubsection{Training Details}
The models is trained by using Adam optimizer with learning rate of $0.00128$ \citep{Kingma2014}. The batch size is set to $10$. The learning rate is halved every 4 epochs and the model is trained until convergence. The input and output images are not normalized and directly used as 8-bits.

Following the common practice, the models are trained until convergence using the scale invariant mean squared error (SMSE). Let $\hat{J}$ be the ground-truth intrinsic image and $J$ be the estimation of the network. Then, mean squared error (MSE) is defined as:

\begin{equation} \label{eq:MSE}
{MSE}(\hat{J}, J) = \frac{1}{N} \sum_{x,y,c}^{} ||\hat{J} - J||^{2}_{2}\;,
\end{equation}

\noindent where $c$ is a color channel index, $x, y$ denotes the pixel coordinate, and $N$ is the total number of valid pixels. Then, SMSE first scales $J$ and then compares its MSE with $\hat{J}$ by least squares:

\begin{equation} \label{eq:SMSE}
{SMSE}(\hat{J}, J) = {MSE}(\alpha J, \hat{J})\;,
\end{equation}
\begin{equation} \label{eq:alpha}
\alpha =  argmin \; {MSE}(\alpha J, \hat{J})\;.
\end{equation}

\noindent Then, MSE and SMSE are combined into one loss function; $\mathcal{L}_{c}$. Thus, to evaluate the quality of the estimation of an intrinsic component, the loss becomes:

\begin{equation} \label{eq:CL}
\begin{aligned}
\mathcal{L}_{c}(J, \hat{J}) = \gamma_{SMSE} \; {SMSE}(\hat{J}, J) + \gamma_{MSE} \; {MSE}(\hat{J}, J)\;,
\end{aligned}
\end{equation}

\noindent where the $\gamma$s are the loss weights. For the experiments, we follow the backbone implementation and set $\gamma_{SMSE}$ to 0.95 and $\gamma_{MSE}$ to 0.05. This is also the common practice in the field. Then, one $\mathcal{L}_{c}$ is assigned to each intrinsic component, yielding four distinct loss functions (reflectances are combined into one, shading, ambient light and shadow cast). Furthermore, an image formation loss (IMF) is included to constrain that the generated final reflectance map and the shading map should follow the image formation model to reconstruct the original image. Enforcing Equation~\eqref{eq:iid}, it compares the input image \emph{I} with the reconstructed image of the predicted reflectance \emph{$\rho$} and shading images ($s_{u}$):
\begin{equation} \label{eq:IMF}
{IMF}(\rho, s_{u}, I) = MSE((\rho \times s_{u}), \;I)\;.
\end{equation}
\noindent Finally, we use the direct shading component as a indirect supervision signal (IS). In Section~\ref{sec:fine_ifm}, we derived the intensity of the composite lighting effects as $s = e_d + e^{+}_a + e^{-}_a$. Since the model predicts ambient light ($e^{+}_a$), shadow cast ($e^{-}_a$) and unified shading ($s_u$) maps, we can reconstruct the direct shading component as follows:
\begin{equation} \label{eq:shading_formation}
e_d = s_u - e^{+}_a - e^{-}_a\;.
\end{equation}
Then, similar to the image formation loss, we compare the calculated direct shading with the ground-truth direct shading images using MSE. Finally, for the generation module, all loss functions are combined as follows:
\begin{equation} \label{eq:loss_functions}
\mathcal{L}_{generator} = \mathcal{L}_{\rho} + \mathcal{L}_{e^{+}_a} + \mathcal{L}_{e^{-}_a} + \mathcal{L}_{IMF} + \mathcal{L}_{IS}\;,
\end{equation}
\noindent where $\mathcal{L}_{\rho}$ is weighted with 1/3, since the loss is defined for three albedo predictions, and $\mathcal{L}_{IMF}$ is weighted with 0.01 and $\mathcal{L}_{IS}$ is weighted with 0.1. Additionally, for the refinement module's final reflectance prediction, supervised L1 pixel loss and L2 gradient losses are utilized. Image gradients are computed by using the intermediate differences between the neighboring pixels. The process is applied vertically and horizontally. Then, L1 pixel loss and L2 gradient losses are added to the generation loss without any weighting. The entire model is trained end-to-end from scratch.

\subsection{Baselines}
\label{sec:baselines}
\begin{figure}[t]
\includegraphics[width=1\linewidth]{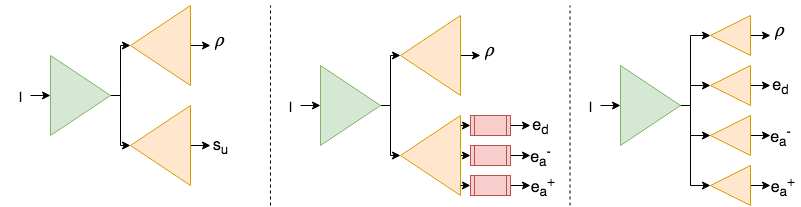}
\centering
\caption{Baseline fine-grained shading models. On the left, a standard encoder-decoder architecture (Eq.~\eqref{eq:iid}), in the middle, \emph{Baseline-a} with $Squeeze-and-Excitation$ blocks \citep{Hu2018}, on the right, \emph{Baseline-b} with extra decoders. $e_d$ denotes direct shading, $e^{-}_a$ is for shadow casts and $e^{+}_a$ is for ambient light, $s_{u}$ is for unified shading, and $\rho$ is for reflectance.}
\label{fig:baseline}
\end{figure}

Since we are the first to estimate fine-grained shading intrinsics, we extend two versions of a state-of-the-art model to provide a fair comparison. The modifications can be applied to any regular encoder-decoder type CNN architecture that is designed for the standard intrinsic image decomposition task. To this end, we extend the ShapeNet model \citep{Shi2017}. The network is designed to further enforce correlations between intrinsic components. Since we increase the number of intrinsic components to predict, the model is well equipped. Figure~\ref{fig:baseline} illustrates the modifications.

For the first baseline, we extend the shading decoder to contain multiple outputs for the photometric effects (\emph{Baseline-a}). The shading decoder includes all shading features which can be further decomposed into the corresponding photometric effects and the direct shading component. To this end, three \emph{Squeeze-and-Excitation} blocks \citep{Hu2018} are added to the end of the shading decoder to perform feature re-calibration. By using SE blocks, predictions of the fine-grained shadings are conditioned by one unified shading decoder enhancing the feature discriminability.

For the second baseline, we extend the main architecture by adding extra decoder blocks per fine-grained shading component (\emph{Baseline-b}). As a result, the architecture has one encoder and four distinct decoders for reflectance, direct shading, shadow cast and ambient light predictions. Unlike \emph{baseline-a}, shading features are not derived from one decoder. Similar to the ShapeNet model, all decoder features are interconnected. In this way, the gradient flow from separate decoders individually increases the feature discriminability.~\footnote{The details of the baseline networks and the training parameters are provided in Appendix~\ref{app:baseline_training_details}}

\section{Experiments and Evaluation}
\label{sec:experiments}
In order to fairly evaluate the capability of the deep intrinsic image decomposition models, along with the baselines, we train several state-of-the-art deep supervised CNN architectures on NED's training split until convergence by using the training details provided by the authors. In addition, we compare our model to a number of state-of-the-art deep unsupervised CNN models. Finally, we include a learning-free optimization based advanced Retinex model \citep{Xu2020}. All the models are directly applied to the test images without any fine-tuning or domain adaptation steps. Following the common practice, we report on the scale-invariant mean squared error (SMSE), where the absolute brightness of each image is adjusted by least squares, the local mean squared error (LMSE) with window size 20, and the structural dissimilarity index (DSSIM) for perceptual quality comparison \citep{Chen2013}. For IIW, Weighted Human Disagreement Rate (WHDR) is provided \citep{Bell2014}.

\subsection{Models}
\noindent\textbf{Direct Intrinsics \citep{Narihia2015}.} It is the first work that directly regresses reflectance and shading maps from an $RGB$ image in a supervised fashion. It adapts a multi-scale architecture that first extracts global contextual information that is then refined using a sub-network. It is trained using SMSE on reflectance and shading predictions and L2 gradient losses on reflectance predictions.
\newline \newline
\noindent\textbf{ShapeNet \citep{Shi2017}.} The model is supervised and it composed of 5 encoder and 5 decoder layers. It exploits the correlations between the intrinsic components by interconnecting decoder features. Skip connections are included from the encoder to the decoders. It is trained using SMSE on reflectance and shading predictions. SMSEs are re-weighted with image gradients for more accurate and sharp predictions.
\newline \newline
\noindent\textbf{IntrinsicNet \citep{Baslamisli2018CVPR}.} The model is supervised and it employs deep VGG16 architectures as encoder and decoders. Skip connections are applied from the encoder to the decoders. It is trained using standard MSE on reflectance and shading predictions together with an image formation loss.
\newline \newline
\noindent\textbf{ParCNN \citep{Yuan2019}.} The model is supervised and it diverges using two distinct encoders, designed as two parallel variant U-Nets. It is trained using SMSE, L1 gradient on reflectance and shading predictions and an image formation loss.
\newline \newline

\noindent\textbf{USI3D \citep{Liu2020}.} The model adversarially learns the latent feature representations of reflectance and shading intrinsics from unsupervised and uncorrelated data. It is trained based on the assumptions about the distributions of different image domains such as domain invariant content, reflectance-shading independence and the reversible latent code encoders.
\newline \newline
\noindent\textbf{IIDWW \citep{Li2018CVPR}} It is an unsupervised deep CNN model trained on image sequences of the same scenes under changing illumination. Therefore, it learns that in a sequence the constant factor is the reflectance and the illumination varies over time. The learning process is further boosted by smoothness priors on both reflectance and shading intrinsics. The evaluations are of particular importance as the model is expected to have strong shadow handling capability due to multi-view exposure. 
\newline \newline
\noindent\textbf{InverseRenderNet \citep{Yu2019}} It is an inverse rendering architecture that is trained using large scale uncontrolled real world outdoor image collections without ground-truths. The network uses 15 encoder layers and 15 decoder layers. It is trained by self supervision modelled by a differentiable renderer and structure-from-motion followed by multi view stereo. Finally, it utilizes a prior constraint on the reflectance generations forcing them to be piecewise smooth. The model appears to be powerful being aware of multi view images, surface normals and a natural illumination prior based on spherical harmonics. 
\newline \newline
\noindent\textbf{STAR \citep{Xu2020}.} It is a structure and texture aware advanced Retinex model based on exponentiated local derivatives. The model is optimization based and does not require labeled data.

 \begin{table*}[h]
 \caption {Full evaluations on NED and MPI Sintel datasets. The baselines improve the ShapeNet backbone. Our model is significantly better than the baselines having fine-grained shadings, state-of-the-art models predicting a unified shading, and the advanced Retinex method. Additionally, our method predicts fine-grained shading components together with the standard intrinsic images. It achieves not only better results but is also more representative. Further, it achieves better generalization performance on MPI Sintel.}
   \scalebox{0.83}{
   \begin{tabular}{|l||c|c|c|c|c|c|||c|c|c|c|c|c|}\hline
      \multirow{3}{*}{} &
       \multicolumn{2}{c|}{SMSE - NED}  &
       \multicolumn{2}{c|}{LMSE - NED} &
       \multicolumn{2}{c|||}{DSSIM - NED} & 
       \multicolumn{2}{c|}{SMSE - Sintel}  &
       \multicolumn{2}{c|}{LMSE - Sintel} &
       \multicolumn{2}{c|}{DSSIM - Sintel} \\
       & Albedo & Shading & Albedo & Shading & Albedo & Shading & Albedo & Shading & Albedo & Shading & Albedo & Shading\\ \hline
      STAR & 0.0174 & 0.0134 & 0.0512 & 0.0486 & 0.4927 & 0.2351 & 0.0242 & 0.0279 & 0.0588 & 0.0610 & 0.3020 & 0.2646\\ \hline
      USI3D & 0.0081 & 0.0143 & 0.0360 & 0.0608 & 0.1886 & 0.2140 & 0.0212 & 0.0304 & 0.0507 & 0.0656 & 0.2688 & 0.2335\\ \hline
      IIDWW & 0.0149 & 0.0175 & 0.0447 & 0.0698 & 0.2229 & 0.2346 & 0.0216 & 0.0273 & 0.0515 & 0.0678 & 0.2672 & 0.2612\\ \hline
      InverseRenderNet & 0.0478 & 0.0505 & 0.0642 & 0.2597 & 0.2751 & 0.3382 & 0.0388 & 0.0446 & 0.0578 & 0.1132 & 0.3069 & 0.2797\\ \hline \hline
      DirectIntrinsics & 0.0089 & 0.0120 & 0.0412 & 0.0680 & 0.2116 & 0.2408 & 0.0257 & 0.0322 & 0.0645 & 0.0780 & 0.3255 & 0.2890\\ \hline
      ShapeNet & 0.0075 & 0.0079 & 0.0276 & 0.0338 & 0.1216 & 0.1176 & 0.0243 & 0.0329 & 0.0562 & 0.0726 & 0.2258 & 0.2071\\ \hline
      IntrinsicNet & 0.0114 & 0.0138 & 0.0333 & 0.0503 & 0.3707 & 0.4583 & 0.0248 & 0.0320 & 0.0546 & \textbf{0.0600} & 0.2077 & 0.2165\\ \hline
      ParCNN & 0.0045 & 0.0052 & 0.0197 & 0.0272 & 0.1129 & 0.0952 & 0.0210 & 0.0271 & 0.0461 & 0.0723 & 0.2251 & 0.1902\\ \hline \hline
      Baseline-a & 0.0072 & 0.0082 & 0.0259 & 0.0387 & 0.1159 & 0.1266 & 0.0233 & 0.0366 & 0.0561 & 0.0708 & 0.2396 & 0.2316\\ \hline
      Baseline-b & 0.0075 & 0.0084 & 0.0280 & 0.0385 & 0.1192 & 0.1340 & 0.0217 & 0.0323 & 0.0519 & 0.0666 & 0.2390 & 0.2214\\ \hline \hline
      ShadingNet (Ours) & \textbf{0.0027} & \textbf{0.0037} & \textbf{0.0122} & \textbf{0.0212} & \textbf{0.0798} & \textbf{0.0788} & \textbf{0.0199} & \textbf{0.0249} & \textbf{0.0448} & 0.0683 & \textbf{0.1991} & \textbf{0.1896}\\ \hline
   \end{tabular}}
   \label{tab:ned_sintel}
  \end{table*}

\subsection{Datasets}
Evaluations are provided on seven datasets. Quantitative results are presented when ground-truth labels are available.
\newline \newline
\noindent\textbf{Natural Environments Dataset (NED).} The test split includes 4800 $RGB$ images with corresponding ground-truths (test data unseen during training). 
\newline \newline
\noindent\textbf{MPI Sintel.} Cross dataset evaluations are provided on the full image set of 890 $RGB$ images with ground-truths. The scenes are rendered from an animated cartoon like short film \citep{Butler2012}. Blender software is used for rendering, but the process, color distributions, and surface, material and camera properties are different from the ones of NED.  
\newline \newline
\noindent\textbf{GTA V.} Cross dataset evaluations are provided on a subset of the test set. The original test set includes scenes having different weather conditions. We exclude scenes with rain, snow, fog as well as night time scenes as they are outside of the scope of this work. Then, we randomly pick 11 scenes yielding around 1800 $RGB$ images. The dataset provides reflectance ground-truths. The scenes are extracted from the Grand Theft Auto V game \citep{Krahenbuhl2018}. The scenes are rendered by a special game engine called Rockstar Advanced Game Engine (RAGE). The rendering process, color distributions, and surface, material and camera properties are different from the ones of NED and MPI Sintel.  
\newline \newline
\noindent\textbf{Intrinsic Images in the Wild (IIW).} Cross dataset evaluations are provided on real world complex scenes. The dataset consists of sparse, crowd-sourced (noisy) relative reflectance annotations on real, mostly indoor images. The annotations are based on the classification task of deciding, given two pair of pixels, which pixel has a darker surface color. From the dataset, we identify around 130 outdoor scenes for evaluations. The images are gathered from Flickr taken from different cameras and setups. See the original work for detailed information on the annotations and the evaluation metric \citep{Bell2014}.
\newline \newline
\noindent\textbf{MIT Intrinsic Images.} Cross dataset evaluations are provided on the real world object-centered images. The dataset initially contains 20 objects. However, we follow the authors' recommendation and exclude apple, pear, phone and potato objects as they are deemed problematic \citep{Grosse2009}. Different from the previous datasets, MIT Intrinsic Images are \emph{object centered} and recorded in a controlled laboratory environment. 
\newline \newline
\noindent\textbf{3DRMS.} Cross dataset evaluations are provided on a real world outdoor garden dataset. The images are recorded by a gardening robot driving through a semantically rich garden with photometric effects \citep{Sattler2017}. The camera setup and the scene properties are similar to the ones of NED. Thus, it can be considered as the real world equivalent of NED. Only qualitative evaluations are provided as the dataset does not provide any ground-truth. 
\newline \newline
\noindent\textbf{Shadow Removal Dataset (SRD).} Cross dataset evaluations are provided on a real world outdoor dataset that is specifically constructed for the shadow removal task. The images were taken by a Canon 5D camera with a tripod, where the shadows are introduced by various objects. It provides a different camera elevation setup. It also includes different illumination conditions, semantically rich scenes, objects with different reflectance phenomena and various shadow silhouettes \citep{Qu2017}. Only qualitative evaluations are provided as the dataset does not provide any ground-truth.

Unfortunately, it is not possible (yet) to densely annotate intrinsic images for any real world outdoor scene. With the current technology, collecting and generating ground-truth real world (object-level) intrinsic images is only possible in a fully-controlled (indoor) laboratory settings \citep{Cheng2019,Grosse2009}. Scene-level densely labeled ground-truth intrinsic images do not exist at all.

 \begin{figure*}[!h]
\includegraphics[width=1\linewidth]{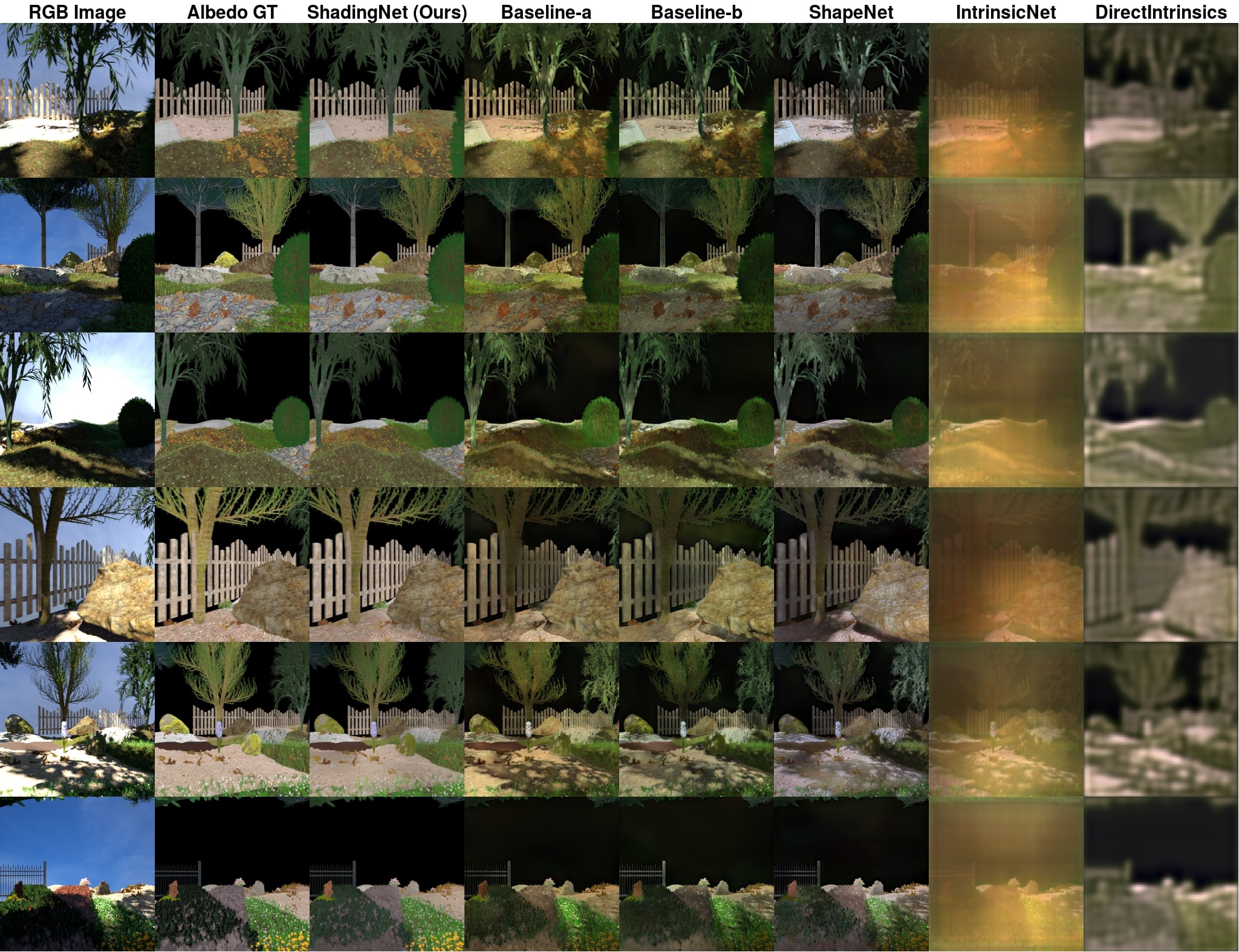}
\centering
\caption{Qualitative reflectance estimation results on NED's test set. Our proposed ShadingNet produces significantly better reflectance images with almost no/minimal shadow leakages, very close to the ground-truth images. A significant visual difference is not observed with the extended baselines and ShapeNet backbone. IntrinsicNet estimations are problematic with undesired color cast artefacts. Similarly, DirectIntrinsics generations appear too blurry and lack proper color information. Images are best viewed in color and on the electronic version.}
\label{fig:ned_albedo}
\end{figure*}

 \begin{figure*}[!h]
\includegraphics[width=1\linewidth]{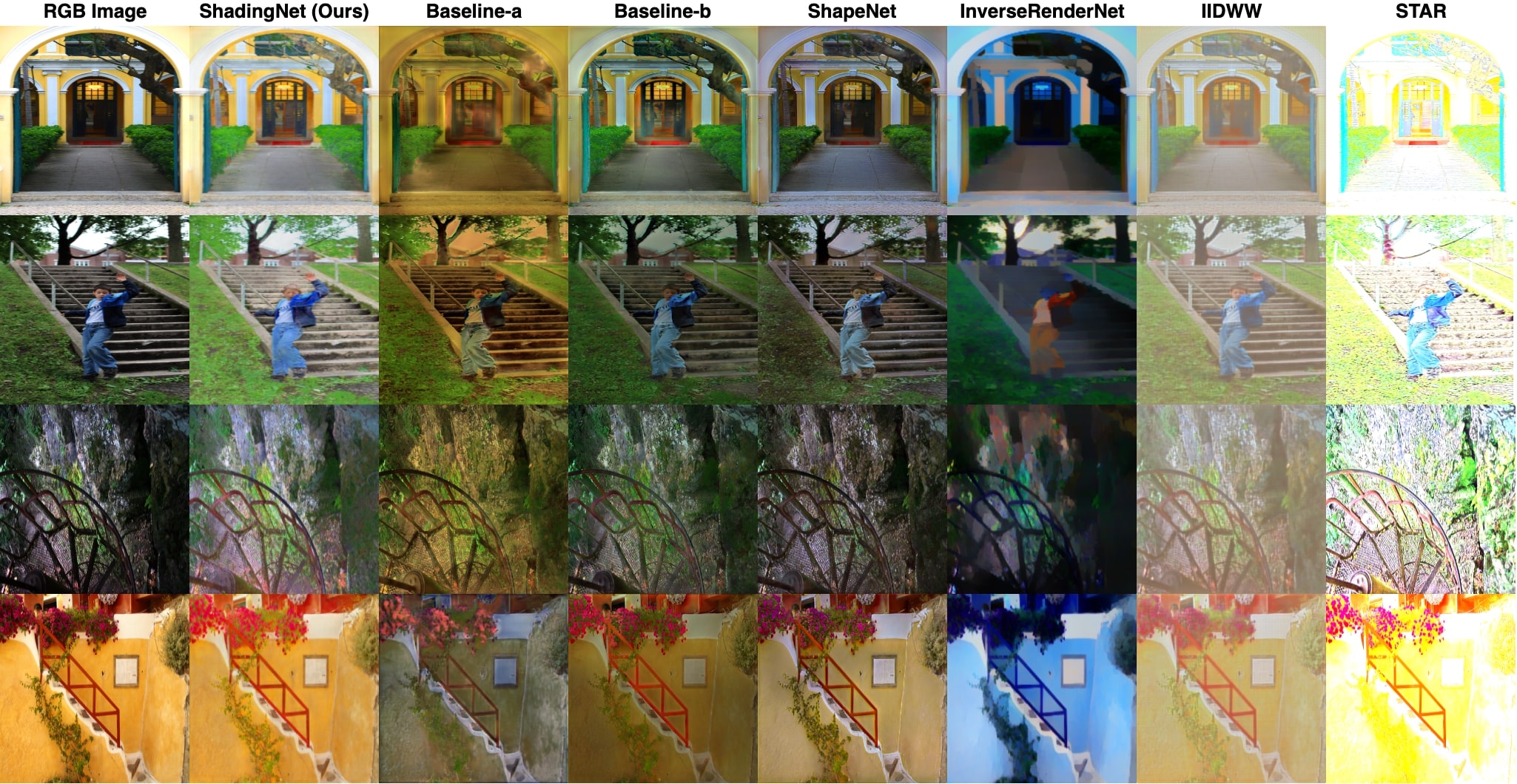}
\centering
\caption{Qualitative estimation results on IIW. Our proposed ShadingNet generated reflectance maps perform better on shadow handling (1st row), strong shading handling (2nd and 4th rows) and low light environments (3rd row). The colors appear more natural and vivid, and the structures are well-preserved. The chromaticity patterns are clearly visible in our reflectance estimations. Best viewed on the electronic version.}
\label{fig:iiw}
\end{figure*}

\subsection{Evaluations on NED, MPI Sintel and GTA V}
\label{evaluations_a}
In this section, we provide extensive quantitative evaluations on three (synthetic) outdoor datasets having completely different rendering processes.  The results are provided in Table~\ref{tab:ned_sintel} for NED and MPI Sintel for full evaluations. Table~\ref{tab:gta} presents reflectance evaluations for GTA V. In addition, Figure~\ref{fig:ned_albedo} provides a qualitative comparison on reflectance estimations, and a qualitative comparison on shading estimations for a number of images with strong photometric effects of NED is provided in Appendix~\ref{app:ned_visuals}. Finally, the qualitative evaluations on MPI Sintel are provided in Appendix~\ref{app:sintel_visuals}.

The quantitative evaluations on the NED's test set (when there is no domain gap) show that the baselines improve the reflectance estimations of the ShapeNet backbone. Hence, further decompositions of the shading component appears to improve reflectance maps by providing explicit photometric cues. A significant difference between the baselines is not observed. On the other hand, the baselines are not as good as our ShadingNet. For all metrics of reflectance and shading, our model outperforms the baselines having fine-grained shadings, state-of-the-art models predicting a unified shading, and the advanced Retinex method.

The qualitative comparisons for reflectance estimations on NED's test set show that the proposed ShadingNet produces significantly better reflectance images with almost no/minimal shadow leakages, very close to the ground-truth images. A significant visual difference is not observed with the extended baselines and ShapeNet backbone. The extended baselines do not exhibit proper shadow handling. IntrinsicNet estimations are problematic with undesired color cast artefacts. Similarly, DirectIntrinsics generations are too blurry and lack proper color information. The qualitative comparisons for shading estimations on NED's test set is provided in Appendix~\ref{app:ned_visuals}.

MPI Sintel and GTA V serve as cross dataset evaluations to assess the generalization capabilities of the models. Note that all the models are directly applied to the test images without any fine-tuning or domain adaptation steps. For MPI Sintel, similar to NED, the baselines further improve the reflectance estimations of ShapeNet backbone. On the other hand, the baselines are not as good as our ShadingNet. Our model outperforms others on all metrics except for the LMSE for shading estimations. Nonetheless, our model is specifically designed to improve reflectance estimations. For GTA V, the baselines cannot further improve the reflectance estimations of ShapeNet backbone. On the other hand, our model is again better on all metrics. The qualitative comparisons show that the proposed ShadingNet generates reflectances that are closer to the ground-truth ones with minimal shadow artefacts and performs better on shadow and low-light handling is provided in Appendix~\ref{app:sintel_visuals}.

\begin{table}[h]
   \centering
   \caption {Reflectance evaluations on GTA V scenes. The baselines cannot further improve ShapeNet backbone. Our model outperforms others on all metrics achieving better generalization capability.}
   \scalebox{0.9}{
   \begin{tabular}{|l||c|c|c|}\hline
      & SMSE & LMSE & DSSIM \\ \hline
     STAR & 0.0165 & 0.0767 & 0.3029 \\ \hline
     USI3D & 0.0129 & 0.0676 & 0.2642 \\ \hline
     IIDWW & 0.0146 & 0.0723 & 0.2713 \\ \hline
     InverseRenderNet & 0.0198 & 0.0884 & 0.2837 \\ \hline \hline
     DirectIntrinsics & 0.0146 & 0.0800 & 0.2981  \\ \hline
     ShapeNet & 0.0138 & 0.0603 & 0.1771 \\ \hline
     IntrinsicNet & 0.0128 & 0.0603 & 0.1989 \\ \hline
     ParCNN & 0.0151 & 0.0656 & 0.4331 \\ \hline \hline
     Baseline-a & 0.0145 & 0.0622 & 0.1883 \\ \hline
     Baseline-b & 0.0134 & 0.0612 & 0.1851 \\ \hline \hline
     ShadingNet (Ours) & \textbf{0.0124} & \textbf{0.0590} & \textbf{0.1698} \\ \hline
   \end{tabular}}
    \label{tab:gta}
 \end{table}

To conclude, experiments conducted on three datasets having completely different rendering processes show that our model with fine-grained shading estimations outperform other methods. Our method has also an improved generalization capability. The baselines having fine-grained shading components further improve ShapeNet backbone on NED and MPI Sintel, but not on GTA V, whereas we achieve superior performance on all. That also highlights the importance of our design choices. Qualitative results further prove the quality of our proposed model. We generate reflectance images with almost no/minimal shadow leakages, and decent colors that are very close to the ground-truth images. Similarly, we achieve sharper shading predictions.

\subsection{Evaluations on Intrinsic Images in the Wild (IIW)}
\label{sec:iiw}
In this section, we provide evaluations on the real world IIW dataset \citep{Bell2014}. The quantitative evaluations are provided in Table~\ref{tab:iiw} and Figure~\ref{fig:iiw} provides qualitative examples for reflectance predictions. The quantitative evaluations show that our model with fine-grained photometric estimations outperforms other learning based methods estimating uniform shading, the inverse rendering model, and the optimization based advanced Retinex method. \emph{Baseline-a} improves ShapeNet backbone results by a small margin, but \emph{Baseline-b} further deteriorates the results by a large margin. State-of-the-art models predicting a unified shading map achieve very similar results. On the other hand, our proposed method suppresses also the baseline models that estimate the fine-grained shading components, which highlights the importance of our design choices. Among the models, IIDWW achieves the best results. We attribute this to the models exposure to image sequences and to the smoothness priors that are used to train the model. It is known that the WHDR metric is biased towards piece-wise smooth reflectance predictions \citep{Nestmeyer2017}. On the other hand, our data driven approach exploits NED using the standard reconstruction losses without any piece-wise smoothness prior. NED includes various rough terrains with different textures, scattered grass, bush and flower patterns that are not piece-wise smooth. Therefore, it is expected that the supervised learning models without proper smoothness constraints may fall short on the metric of the dataset. Nonetheless, it is possible to further improve the performance by applying a guided filter \citep{Nestmeyer2017} to enforce the piece-wise constant reflectance assumption as a post-processing step.
\begin{table}[h]
   \centering
   \caption {Reflectance evaluations on real world IIW outdoor scenes. Our method achieves superior results also on real world complex outdoor images. $*$ indicates that the CNN predictions are post-processed with a guided filter \citep{Nestmeyer2017}.}
   \scalebox{0.9}{
   \begin{tabular}{|l||c|}\hline
      & WHDR$~\downarrow$ \\ \hline
     STAR & 36.21  \\ \hline
     USI3D & 36.69  \\ \hline
     IIDWW & \textbf{21.60}  \\ \hline
     InverseRenderNet & 36.05  \\ \hline \hline
     DirectIntrinsics & 41.64  \\ \hline
     ShapeNet & 40.33 \\ \hline
     IntrinsicNet & 38.17 \\ \hline
     ParCNN & 40.06 \\ \hline \hline
     Baseline-a & 46.22 \\ \hline
     Baseline-b & 39.11 \\ \hline \hline
     ShadingNet (Ours) & \underline{35.73} \\ \hline
     ShadingNet (Ours)$^{*}$ & \underline{29.98} \\ \hline
   \end{tabular}}
   \label{tab:iiw}
 \end{table}
 
The merits of the results are more compelling when evaluated visually. The qualitative comparisons show that \emph{Baseline-a} predictions are corrupted by a yellowish color cast and the model fails to generate a proper reflectance image when the scene is dominated by a single color as in the case of the 4th row. \emph{Baseline-b} predictions displaying partial chromaticity information in several regions are relatively better than \emph{Baseline-a} ones, but the generated colors appear rather dull. ShapeNet estimations strongly resembles the input image intensities. Although it is possible to see the smoothing effects, InverseRenderNet predictions fail most of the time. The model fails to properly generate colors as in the cases of the 1st and 4th rows having blue color casts. It further generates artefacts in the 2nd row where the face of the boy appears blue, his blue jeans appear brown and his blue jacket appears dark red. The shading effects are still visible. IIDWW colors are rather dull and faded having blur like casts. We attribute this to the smoothness losses used in its training. Quantitatively it contributes, but qualitatively they generate undesired effects. STAR estimations can handle strong shadings, but appear way too bright that most of the structures and colors are not visible anymore. The model appears to handle low light conditions better than other learning based models. Qualitative comparison with other models are provided in Appendix~\ref{app:iiw_visuals}.

 \begin{table*}[t]
 \centering
 \scalebox{0.9}{
   \begin{tabular}{|l||c|c|c|c|c|c|c|c|c|}\hline
      \multirow{2}{*}{} &
       \multicolumn{3}{c|}{SMSE}  &
       \multicolumn{3}{c|}{LMSE} &
       \multicolumn{3}{c|}{DSSIM} \\
       & Albedo & Shading & Average & Albedo & Shading & Average & Albedo & Shading & Average \\ \hline
      STAR & 0.0137 & 0.0114 & 0.0126 & 0.0614 & 0.0672 & 0.0643 & 0.1196 & \textbf{0.0825} & 0.1011 \\ \hline 
      USI3D & 0.0156 & 0.0102 & 0.0129 & 0.0640 & 0.0474 & 0.0557 & 0.1158 & 0.1310 & 0.1234 \\ \hline 
      IIDWW & 0.0126 & 0.0105 & 0.0116 & 0.0591 & 0.0457 & 0.0524 & 0.1049 & 0.1159 & 0.1104 \\ \hline 
      InverseRenderNet & 0.0234 & 0.0137 & 0.0186 & 0.0573 & 0.0957 & 0.0765 & 0.1148 & 0.1276 & 0.1212 \\ \hline \hline
      DirectIntrinsics & 0.0164 & 0.0093 & 0.0129 & 0.0683 & 0.0449 & 0.0566 & 0.1218 & 0.1159 & 0.1189 \\ \hline
      ShapeNet & 0.0207 & 0.0106 & 0.0157 & 0.0606 & 0.0595 & 0.0601 & 0.1027 & 0.0886 & 0.0957 \\ \hline
      IntrinsicNet & 0.0191 & 0.0089 & 0.0140 & 0.0618 & \textbf{0.0407} & 0.0513 & 0.0905 & 0.0989 & 0.0947 \\ \hline
      ParCNN & 0.0109 & 0.0086 & 0.0098 & 0.0462 & 0.0537 & 0.0500 & 0.0929 & 0.0999 & 0.0964 \\ \hline \hline
      Baseline-a & 0.0141 & 0.0089 & 0.0115 & 0.0523 & 0.0548 & 0.0536 & 0.0929 & 0.0947 & 0.0938 \\ \hline
      Baseline-b & 0.0156 & 0.0086 & 0.0121 & 0.0563 & 0.0522 & 0.0543 & 0.0939 & 0.0953 & 0.0946 \\ \hline \hline
      ShadingNet (Ours) & \textbf{0.0107} & \textbf{0.0071} & \textbf{0.0089} & \textbf{0.0390} & 0.0447 & \textbf{0.0419} & \textbf{0.0758} & 0.0865 & \textbf{0.0812} \\ \hline
   \end{tabular}}
   \caption{The baselines further improves ShapeNet backbone. Our reflectance estimations are significantly better than others also on on object-centered MIT Intrinsic Images demonstrating superior generalization ability.}
   \label{tab:mit}
  \end{table*}

On the other hand, our proposed ShadingNet generates significantly better reflectance maps. Colors appear more natural and vivid, and the structures are well-preserved. The 1st row shows that we perform better on shadow handling as the shadows due to the bushes are mostly eliminated. The stairs in the 2nd row and upper background of the 4th row present strong shading patterns and our model is capable of handling them properly. The jeans of the person in the 2nd row presents abrupt geometry changes and our model is able to generate relatively smooth reflectance predictions in that case. In addition, the 3rd row presents a case of a low light environment that our model presents better handling capability. Finally, the chromaticity patterns are clearly visible in our reflectance predictions: the green bushes and the red carpet in the 1st row, the green grass and the green tree leaves in the 2nd row, the green moss puddle in the 3rd row and the red flowers and the red stair rail in the 4th row. The results are particularly important as the chromaticity information perfectly separates reflectances under ideal conditions. Thus, our model is capable of extracting meaningful reflectance information from the $RGB$ images.

 \begin{figure}[t]
\includegraphics[width=1\linewidth]{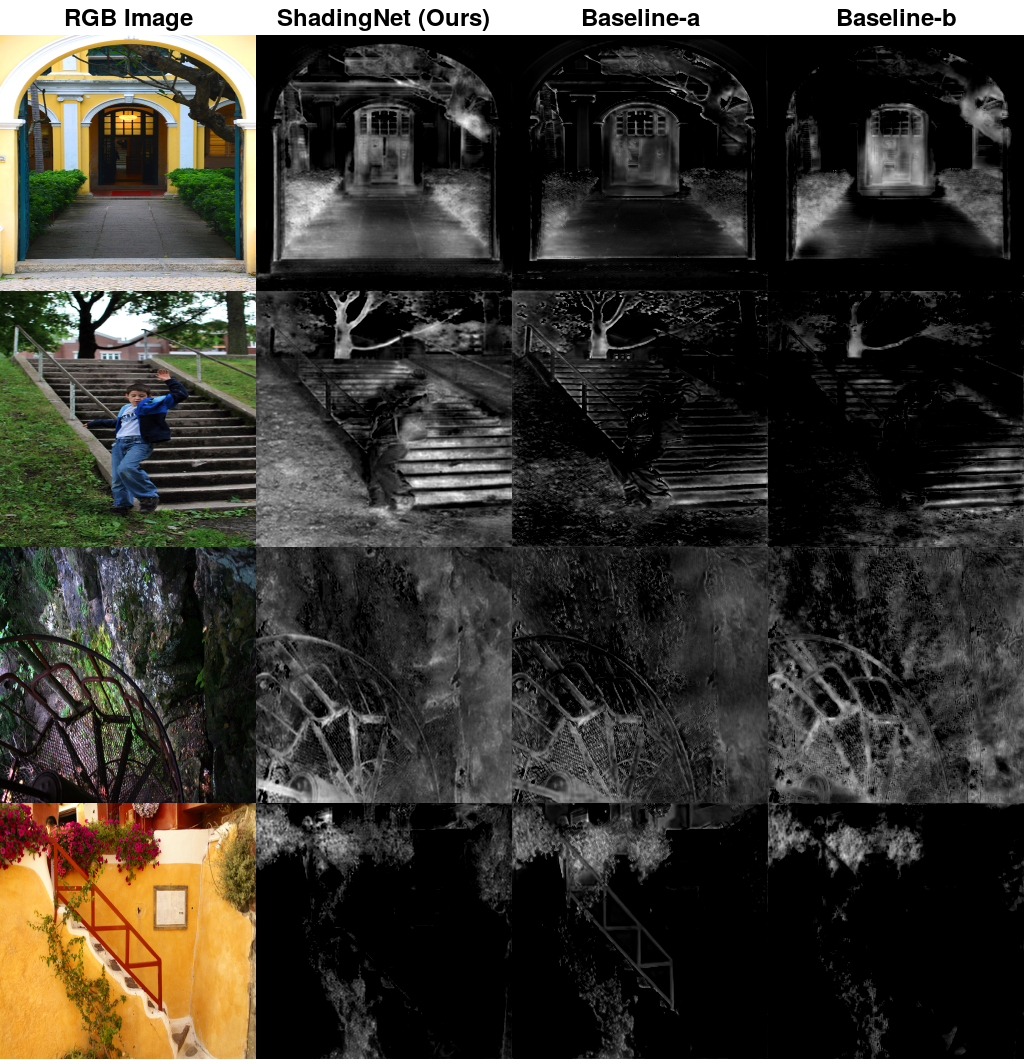}
\centering
\caption{Shadow evaluations on real world IIW scenes. Our model generates sharper and richer shadow maps capable of handling diverse cues for complex scenes.}
\label{fig:iiw_shadow}
\end{figure}

Finally, Figure~\ref{fig:iiw_shadow} provides shadow map evaluations. The results prove that our model generates richer shadow maps. The 1st row shows that our model captures the shadows on the hallway, on the portico, on the left arch, on the eaves and due to the self occlusions of the bushes. Baselines fail to capture the shadows on the hallway, on the left arch, and on the eaves. The 2nd row shows that our model is better at detecting the self occlusions of the grass and tree leaves and the strong intensity drops of the stairs. The 3rd row shows that all models perform similarly on low light conditions, yet our model estimations appear sharper. The 4th row shows that all models are capable of detecting shadows due to the self occlusions of the ivy and the flowers. However, \emph{Baseline-a} wrongly detects the stair rails as shadowy regions and \emph{Baseline-b} fails to detect the region over the upper right background.

To conclude, ShadingNet generated reflectance images emerge more stable, obtain better quantitative results, have more vivid, realistic and natural colors, can handle strong shadings, low light environments and abrupt geometry changes, and have better generalization capability for in-the-wild real world complex outdoor images. Our shadow images appear sharper and they are capable of handling a lot more diverse cues for various complex outdoor scenes. Although IIDWW achieves better quantitative results, it uses images sequences for training and applies smoothness priors to address the metric. We show that explicitly estimating photometric effects further contributes to improve the reflectances.
 
\subsection{Evaluations on MIT Intrinsic Images}
\label{sec:mit}
\begin{figure}[t]
\includegraphics[width=1\linewidth]{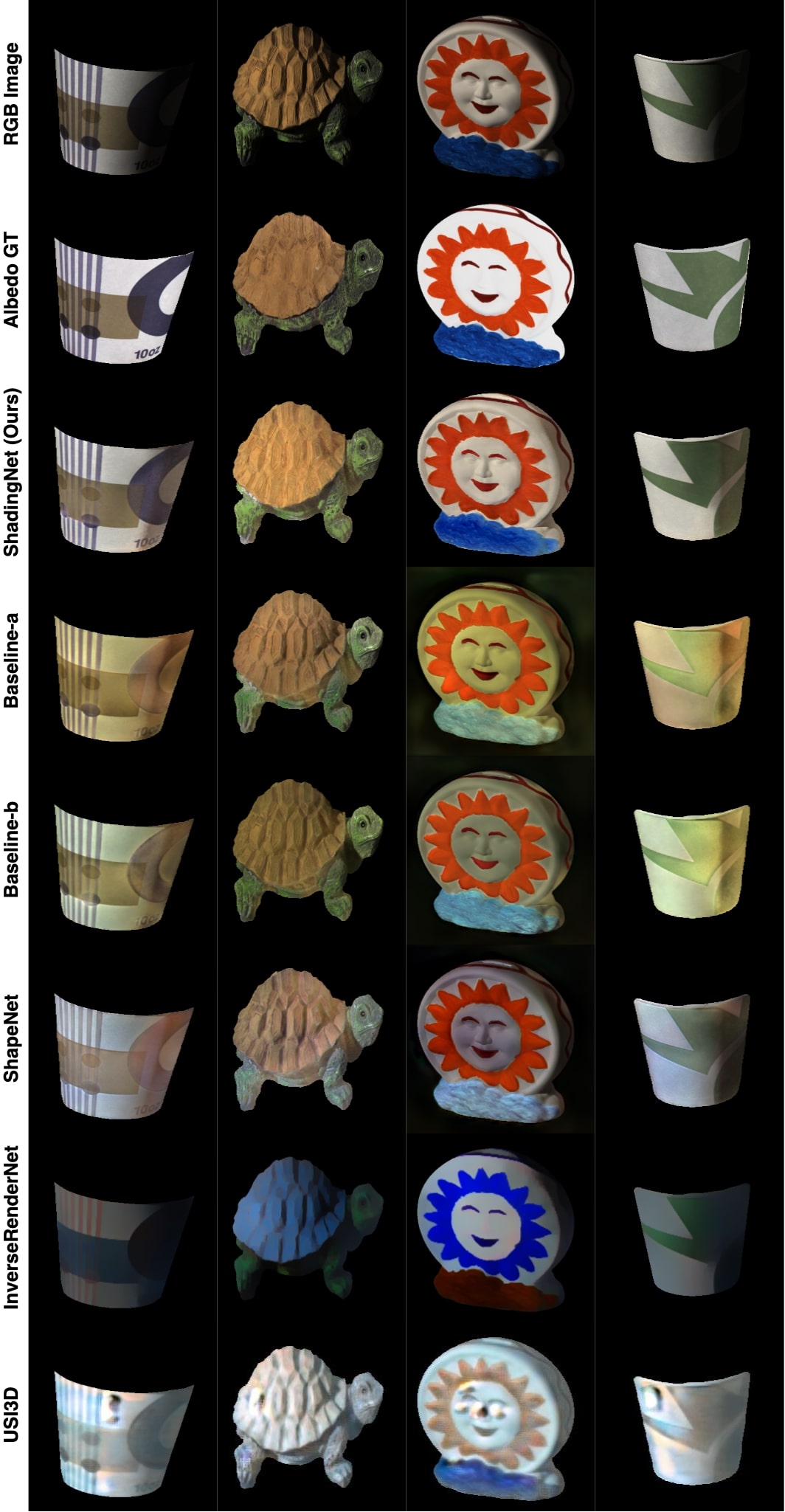}
\centering
\caption{We generate sharper estimations that are closer to the ground-truths with better color reproduction and shading handling. We exhibit significantly better generalization performance also on MIT Intrinsic Images.}
\label{fig:mit}
\end{figure}

In this section, we evaluate our model on the real world object-centered MIT Intrinsic Images \citep{Grosse2009}. The quantitative evaluation results are provided in Table~\ref{tab:mit} and Figure~\ref{fig:mit} provides a number of qualitative examples for reflectance predictions.

The quantitative results demonstrate that the baselines estimating fine-grained shading components further improve the reflectance estimations of ShapeNet backbone estimating a unified shading. On the other hand, our proposed ShadingNet generates reflectance maps that significantly outperforms all other methods on all metrics. IntrinsicNet shading estimations achieve better results for the LMSE for shading estimations and the learning free advanced Retinex model STAR achieves better shading estimations for the DSSIM metric. Nonetheless, on average we achieve significantly better results for all cases. Moreover, our shading estimations appear competitive and our model is specifically designed to improve reflectance estimations and all the reflectance metrics are improved also for a real world object-centered dataset of a completely different domain.

\begin{figure*}[p]
\includegraphics[scale=0.38]{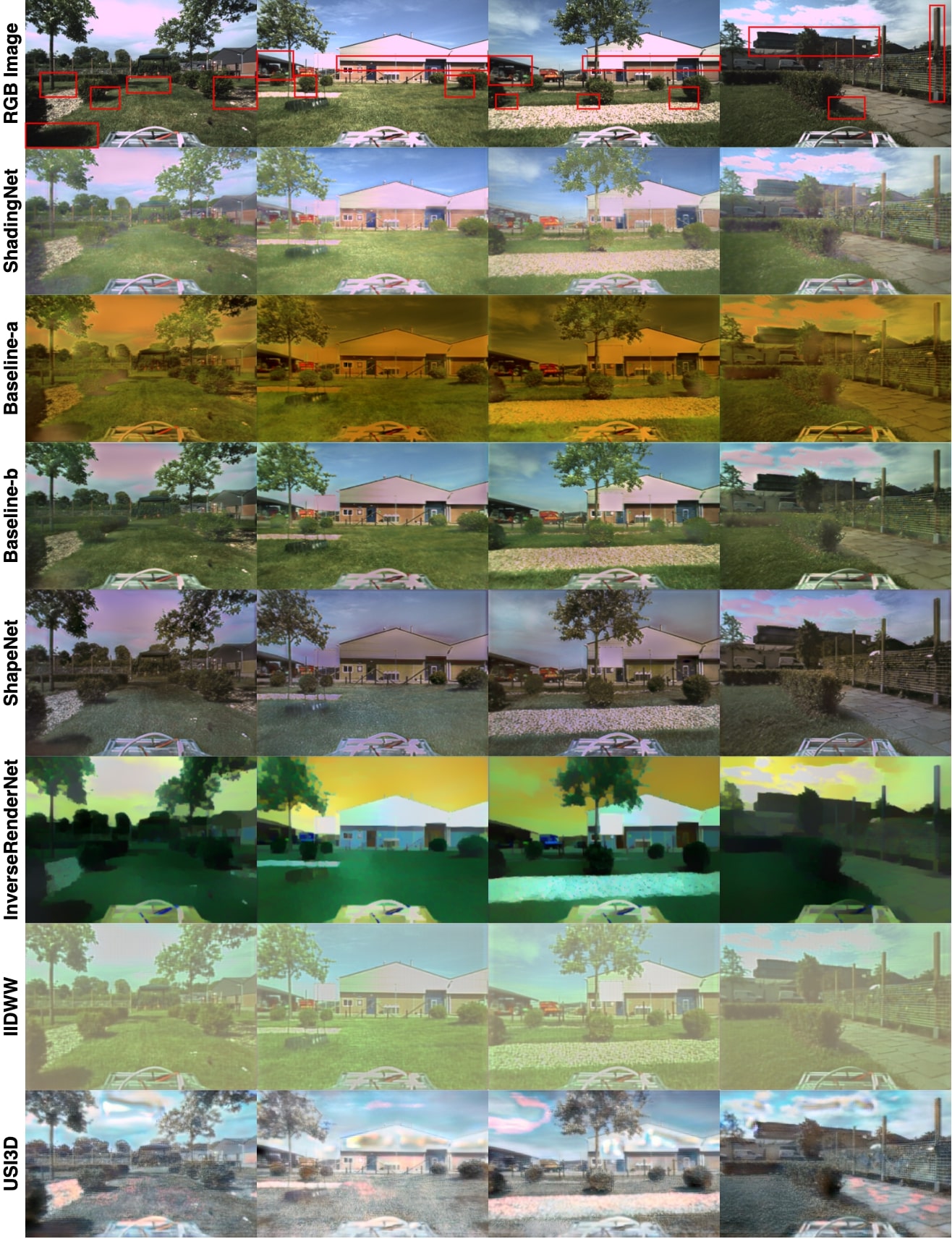}
\centering
\caption{Qualitative reflectance evaluations on a real world garden images. Our proposed ShadingNet generates significantly better reflectance maps that the colors appear more natural and vivid, and the structures are well-preserved. We perform better on handling various shadow patterns and also low light conditions.}
\label{fig:trimbot}
\end{figure*}

\begin{figure*}[h]
\includegraphics[width=1\linewidth]{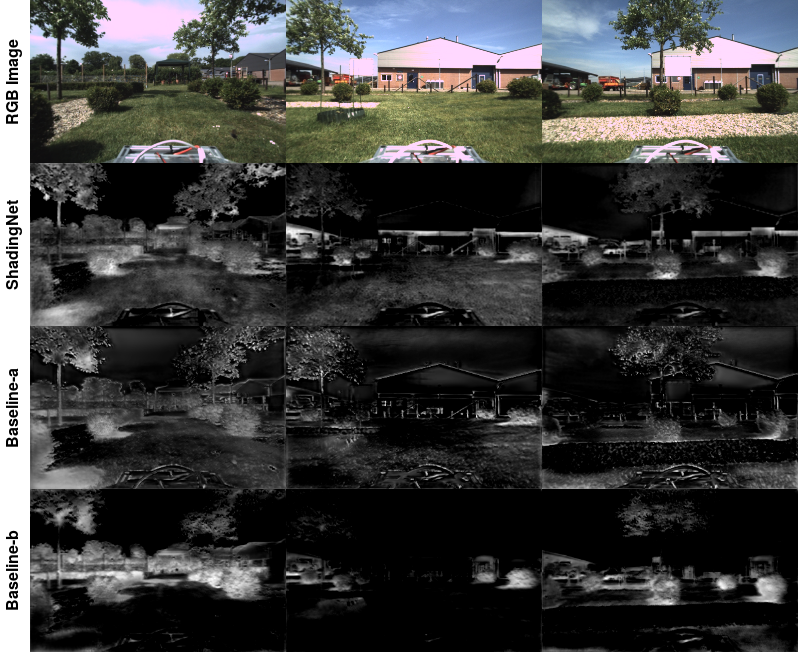}
\centering
\caption{Shadow evaluations on real world garden scenes. \emph{Baseline-a} wrongly focuses to the sky. \emph{Baseline-b} appears to be limited by the field of view and the depth, and it puts too much emphasis on the self occlusions of the bushes rather than the strong shadow casts. We generate richer shadow maps with diverse shadow patterns.}
\label{fig:trimbot_shadow}
\end{figure*}

\begin{figure*}[h]
\includegraphics[width=1\linewidth]{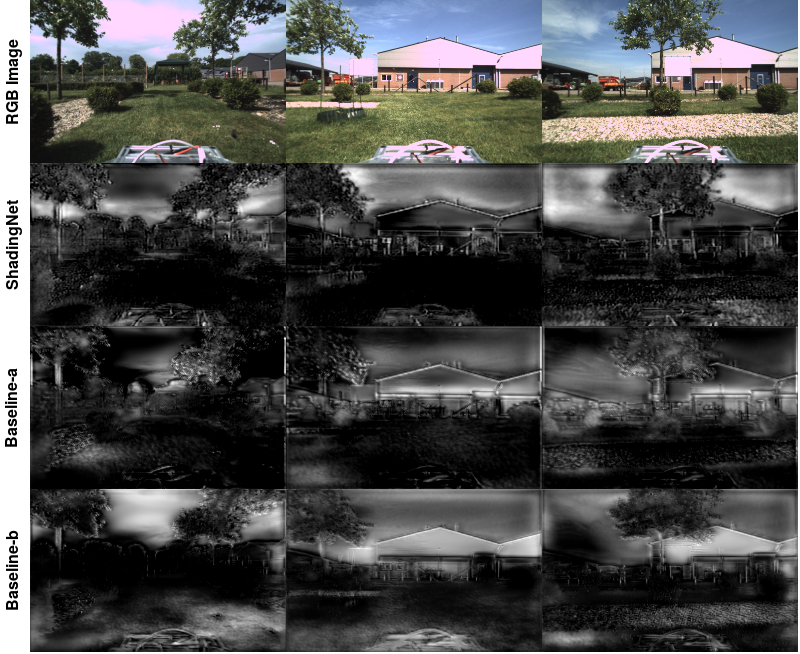}
\centering
\caption{Ambient light evaluations on real world garden scenes. The baselines overfit to the brightest pixels as ambient light predictions. Our model can better differentiate brightness changes and attribute them to reflectance or illumination.}
\label{fig:trimbot_ambient}
\end{figure*}

The qualitative results demonstrate that our model generates better reflectance maps than other learning-based models. Similar to the previous experiments, the baselines tend to generate undesired yellowish color cast. ShapeNet generated colors are rather dull and faded. Similar to IIW experiments, InverseRenderNet tends to confuse color with light generating erroneous reflectance images. Moreover, it fails to handle shading cues from reflectance predictions. USI3D generated reflectance maps are contaminated by artefacts and the colors are rather dull and faded. On the other hand, ShadingNet generated reflectance images are sharper and closer to the ground-truth images with better color reproduction and shading handling. Qualitative comparison with other models are provided in Appendix~\ref{app:mit_visuals}.

Previous experiments on scene-level NED, MPI Sintel, GTA V and IIW have already proved the greater generalization ability of the proposed ShadingNet. In addition to those, ShadingNet model offers significantly better performance also on a totally different domain of real world object-centered images both quantitatively and qualitatively. Therefore, ShadingNet model presents an exceptional generalization performance compared with the baselines having fine-grained shadings, state-of-the-art models predicting a unified shading, and the advanced Retinex method.

\subsection{Evaluations on 3DRMS of Outdoor Garden Scenes}
\label{sec:trimbot}
In this section, we present extensive qualitative comparisons for a real world in-the-wild outdoor garden dataset, 3DRMS \citep{Sattler2017}. It can be considered as the real world equivalent of NED. The results are presented in Figure~\ref{fig:trimbot} for reflectance predictions. 

Similar to the previous experiments, \emph{Baseline-a} predictions are corrupted by a yellowish color cast. The model detects a number of shadow patterns, but it cannot properly eliminate them from the reflectance generations. \emph{Baseline-b} provides better shadow handling than \emph{Baseline-a} and also from ShapeNet backbone. ShapeNet model fails to generate proper colors and wrongly classifies shadows cues into reflectance predictions. InverseRenderNet model generates smooth reflectance predictions. However, it has no sense of shadows or other photometric effects. All the reflection predictions are contaminated by shadow cues. Furthermore, the surfaces with lighter colors are negatively affected by green color casts. Similar behaviour is also observed for IIDWW that the images have green color casts. This may indicate that the models fail when a scene is dominated by a single color (green in this case). Moreover, IIDWW predictions highly resemble the input $RGB$ images. Although the model is trained with image sequences and is expected to have high quality shadow handling, it fails to achieve so. Finally, USI3D estimations are contaminated with artefacts and the colors are rather dull and faded. Similar to InverseRenderNet and IIDWW, the model has no sense of shadows. Qualitative comparison with other models are provided in Appendix~\ref{app:trimbot_visuals}.

On the other hand, our proposed ShadingNet generates better reflectance maps that the colors appear more natural and vivid, and the structures are well-preserved. Especially, the trees appear more lively. The 1st column shows that we perform better on shadow handling as the shadows due to the bushes and the trees are mostly eliminated. Similar behaviour is also observed on 2nd and 3rd columns that the shadows due to to the bushes are handled well. Even the strong shadow cast patters on the eaves of the buildings are mostly eliminated. None of the other models is able to handle or recognize them. Likewise, our model is better at handling the shadows of the boxwood and the shadings of the posts, and the building in front with low light is more visible than others.

Finally, we provide the qualitative evaluations of the photometric effect estimations. Figure~\ref{fig:trimbot_shadow} provides shadow generation evaluations. \emph{Baseline-a} wrongly focuses to numerous regions of the sky. \emph{Baseline-b} appears to be limited by the field of view and the depth as it does not generate any information for further away and small objects. For example, the regions with the background buildings are totally discarded and the self occlusions of tree on the 3rd column is completely neglected. On the other hand, our model can generate richer shadow maps and it is aware of various shadow patterns. The 1st column shows that the baselines cannot capture the shadows due to the bushes placed on the right. \emph{Baseline-b} cannot detect the shadow of the bottom left bush. In addition, it puts too much emphasis on the self occlusions of the bushes rather than the strong shadow casts. 2nd and 3rd columns show that our model can detect the strong shadow casts of the eaves of the background buildings of the right sides, and the very strong shadow cast on the background building of the left side. \emph{Baseline-b} detects the one on the left up to a degree, but the baselines fail to detect the strong shadow casts of the eaves.

Figure~\ref{fig:trimbot_ambient} presents ambient light estimation evaluations. All models mostly focus on the sky as our image formation and data generation process model ambient illumination as the extra light present on top of the direct shading component. Although the training data do not include sky regions (since it is not possible to generate proper synthetic ground-truth for the sky), the models are mostly aware of the nature of the ambient light. However, the baselines directly anticipate that the brightest pixels highlight ambient light cues. It can be observed from the figure that the baselines put much of the emphasis on the aluminum-like roof covers of the buildings with shiny reflectance properties, because these regions are among the most bright pixels of that scenes. Thus, the results suggest that the baselines overfit to the brightest pixels as ambient light predictions. On the other hand, our model can better differentiate brightness changes and attribute them to reflectance or illumination. It can be observed from the 2nd and 3rd columns that our model mostly focuses on the sky rather the shiny roof reflectance, whereas \emph{Baseline-a} highlights more to the roof material and the shadow cast edges and \emph{Baseline-b} uniformly highlights the roof material and puts less emphasis on the sky.

To conclude, ShadingNet generated reflectance images emerge more stable, generate more vivid, realistic and natural colors, can handle diversified shadow patterns and low-light environments, and have better generalization capability for in-the-wild real world outdoor garden scenes. Our shadow estimations are capable of capturing a lot more diverse cues, they are aware of the sky, and they are not limited by the depth of the scene. Finally, our model can better differentiate brightness changes and attribute them to reflectance or illumination, whereas the baselines directly overfit to the brightest pixels.

\begin{figure*}[p]
\includegraphics[scale=0.38]{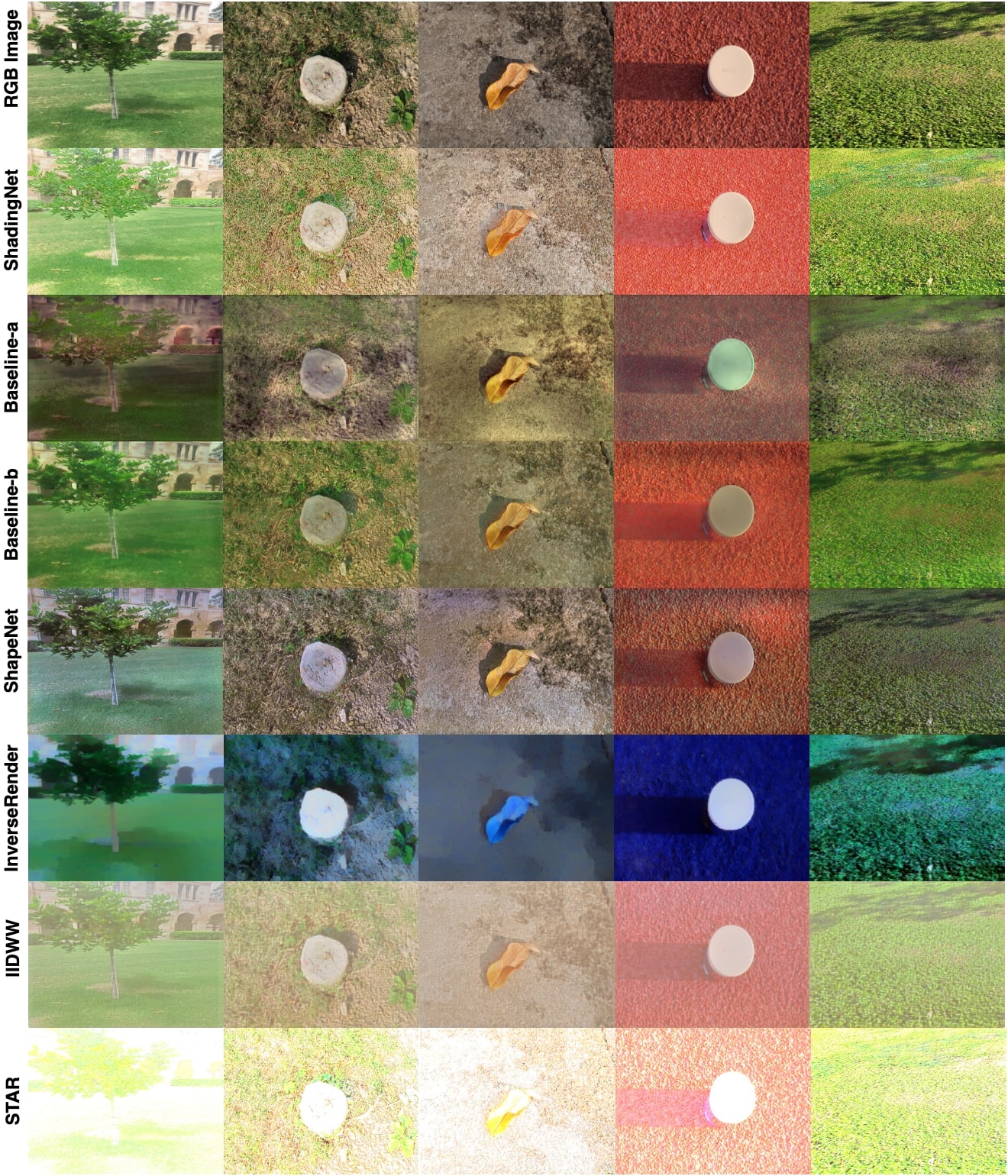}
\centering
\caption{Qualitative reflectance evaluations on a real world shadow removal dataset. Our proposed ShadingNet generates significantly better reflectance maps that the colors appear more natural and vivid, and the structures are well-preserved, whereas other learning-based methods cannot properly handle shadow casts and some even generate additional artefacts. Our model is also able to remove shadow casts on achromatic surfaces (3rd column). Images are best viewed in color and on the electronic version.}
\label{fig:srd}
\end{figure*}

 \begin{figure*}[!h]
\includegraphics[width=1\linewidth]{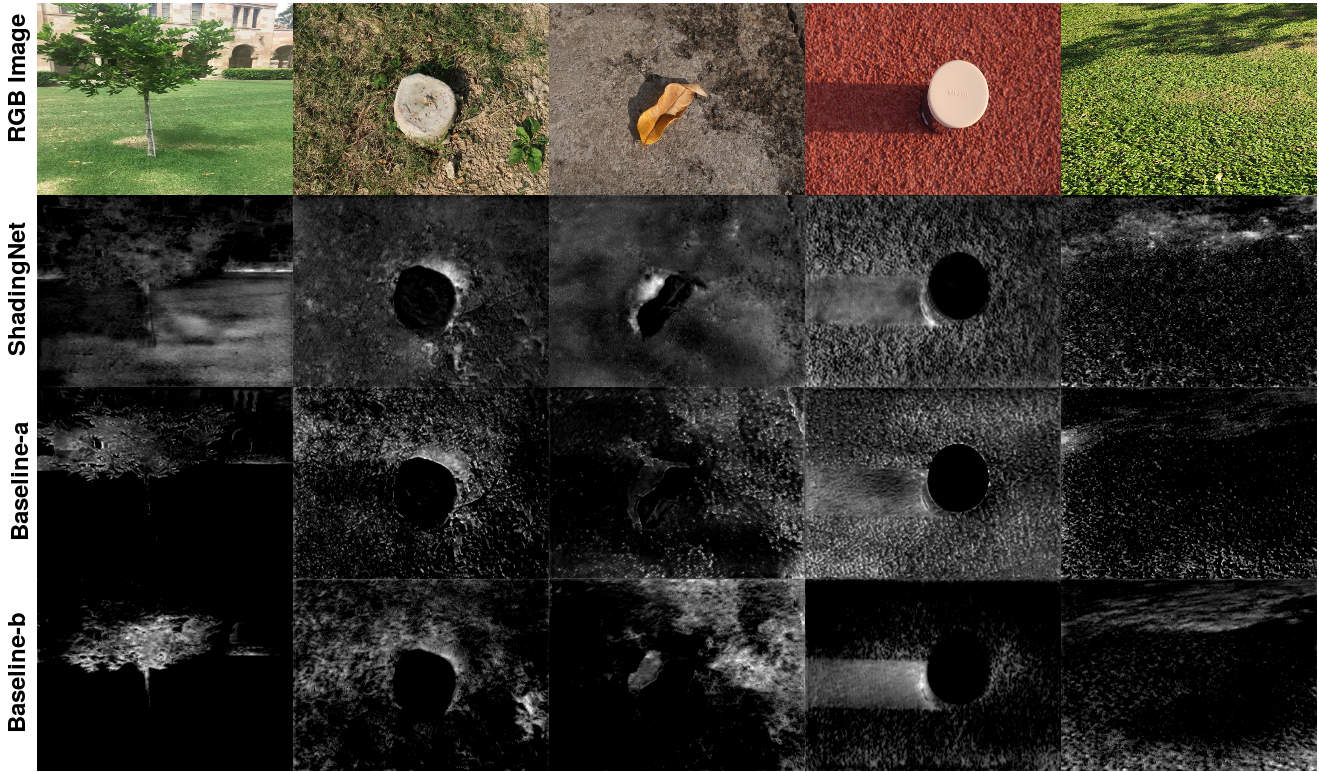}
\centering
\caption{Shadow cast evaluations on a real world shadow removal dataset. Our model can generate richer shadow maps, it can capture various shadow patterns, and it is aware of achromatic surfaces.}
\label{fig:srd_shadow}
\end{figure*}

\subsection{Evaluations on Shadow Removal Dataset (SRD)}
\label{sec:srd}
In this section, we demonstrate the quality of our model on a real world complex outdoor dataset that is specifically crafted for the shadow removal task \citep{Qu2017}. A number of qualitative comparisons are provided in Figure~\ref{fig:srd} for reflectance predictions. 

The results suggest that the baselines do not further improve ShapeNet backbone, yet \emph{Baseline-b} generates better colors. \emph{Baseline-a} generates additional shadow artefacts (1st, 2nd and 5th colums). ShapeNet predictions have rather dull colors. Similar to previous experiments, InverseRenderNet tends to confuse color with light generating erroneous reflectance images. The reflectance estimation in the 4th column is completely off. In addition, instead of removing shadow cues from reflectance images, the model further \emph{boosts} the shadow pixels. For example, in the 1st column, the area under the tree is further clustered and in the 5th column, the intensity and contrast of the shadow is further emphasized. It suggests that the model cannot properly handle outdoor shadows. IIDWW estimations resemble the input $RGB$ images. Similar to the previous experiments, the images are rather dull in color and all the shadow cues are also present in the reflectance estimations. Although the model is trained on image sequences, it cannot properly handle outdoor shadows. All learning-based methods fail on handling shadow casts on SRD. STAR model can handle the shadow casts of 2nd, 3rd and 5th columns. However, for others, it again generates reflectance images that are way too bright that most of the structures and colors are not visible anymore. For example, the background building and the boxwood of the 1st column is not visible anymore. Qualitative comparison with other models are provided in Appendix~\ref{app:srd_visuals}. 

On the other hand, our proposed ShadingNet generates significantly better reflectance maps that the colors appear more natural and vivid, and the structures are well-preserved. The 1st column shows that we perform better on shadow handling as the shadows below the tree and shadows due to the self occlusion of the boxwood are mostly eliminated. Similarly, we can handle the relatively small shadow cast of the stone in the 2nd column. In addition, we are the only model that can properly handle the shadow cast of the leave where the surface is achromatic. In the 4th column, the uniformity of the background is observed with minimum shadow leakage, whereas other models generate additional light artefacts. Finally, the 5th column appears more realistic and natural. 

Finally, Figure~\ref{fig:srd_shadow} provides the qualitative evaluations of the shadow cast estimations. The 1st column shows that the baselines can only detect the self occlusions of the tree leaves and a small part on top right where the boxwood meets the ground. Our model can fully detect the intersection between the boxwood and the ground (also the on the left), the shadow region below the tree and even micro self occlusions of the grass. The models behave practically the same in the 2nd column. The 3rd column shows that our model can fully detect the shadow cast of the leave on an achromatic surface, whereas the baselines focus more on the region where the darker pixels are distributed. \emph{Baseline-b} can partially detect it, but most of the emphasis is on the darker brownish moss puddle which should have been attributed to the reflection estimation. In the 4th column, \emph{Baseline-b} generates relatively sharper estimations, yet it ignores the rugged nature of the surface of the running track which cause small shadows due to self occlusions. The 5th column shows our model can detect the shadow cast better, whereas \emph{Baseline-a} fails.

To conclude, ShadingNet generates significantly better reflectance intrinsics also on a different camera elevation setup. Our generated colors appear more natural and vivid. In addition, our model can handle various shadow cast patterns including achromatic surfaces and single color dominated scenes, whereas the baselines generally struggle to cover diverse patterns and distributions.

\subsection{Evaluation of the Refinement Module}
\label{sec:refinement_ablation}
In this section, we evaluate the quality of the individual reflectance estimations and the final refinement module. Quantitative results are provided in Table~\ref{tab:refinement} for NED when there is no domain gap and for GTA V as cross dataset evaluation to assess the generalization performance. It can be observed that the refinement module further improves the reconstruction quality of the reflectance estimations for all metrics. The reflectance maps estimated from the shadow branch achieve the best results compared with other photometric effects. It suggests that the strong shadow cast cues negatively effect reflectance estimations the most. Explicitly classifying them generates better reflectance maps. 

\begin{table}[h]
\centering
\caption{Evaluation of the refinement module. It further improves the reconstruction quality of the reflectance maps for all metrics. $\rho_{a^{+}}$ denotes the ambient light branch, $\rho_{a^{-}}$ denotes the shadow cast branch, $\rho_{u}$ denotes the unified shading branch reflection predictions. $\rho$ denotes the final refined reflectance estimation.}
\begin{subtable}{1\columnwidth}
\centering
\caption{Evaluations on NED (when there is no domain gap).}\label{tab:sub_first}
   \begin{tabular}{|l||c|c|c|}\hline
       & SMSE & LMSE & DSSIM \\\hline
      $\rho_{a^{+}}$ & 0.0032 & 0.0160 & 0.1005 \\ \hline
      $\rho_{a^{-}}$ & 0.0030 & 0.0144 & 0.0910 \\ \hline
      $\rho_{u}$ & 0.0030 & 0.0157 & 0.0982 \\ \hline \hline
      $\rho$ & \textbf{0.0027} & \textbf{0.0122} & \textbf{0.0798} \\\hline
   \end{tabular}
\end{subtable}

\bigskip
\begin{subtable}{1\columnwidth}
\centering
   \caption{Evaluations on GTA V (as cross dataset generalization).}\label{tab:sub_second}
   \begin{tabular}{|l||c|c|c|}\hline
       & SMSE & LMSE & DSSIM \\\hline
      $\rho_{a^{+}}$ & 0.0130 & 0.0620 & 0.2169 \\ \hline
      $\rho_{a^{-}}$ & 0.0125 & 0.0598 & 0.1970 \\ \hline
      $\rho_{u}$ & 0.0127 & 0.0613 & 0.2096 \\ \hline \hline
      $\rho$ & \textbf{0.0124} & \textbf{0.0590} & \textbf{0.1968} \\\hline
   \end{tabular}
\end{subtable}
\label{tab:refinement}
\end{table}

\begin{figure}[h]
\includegraphics[width=1\linewidth]{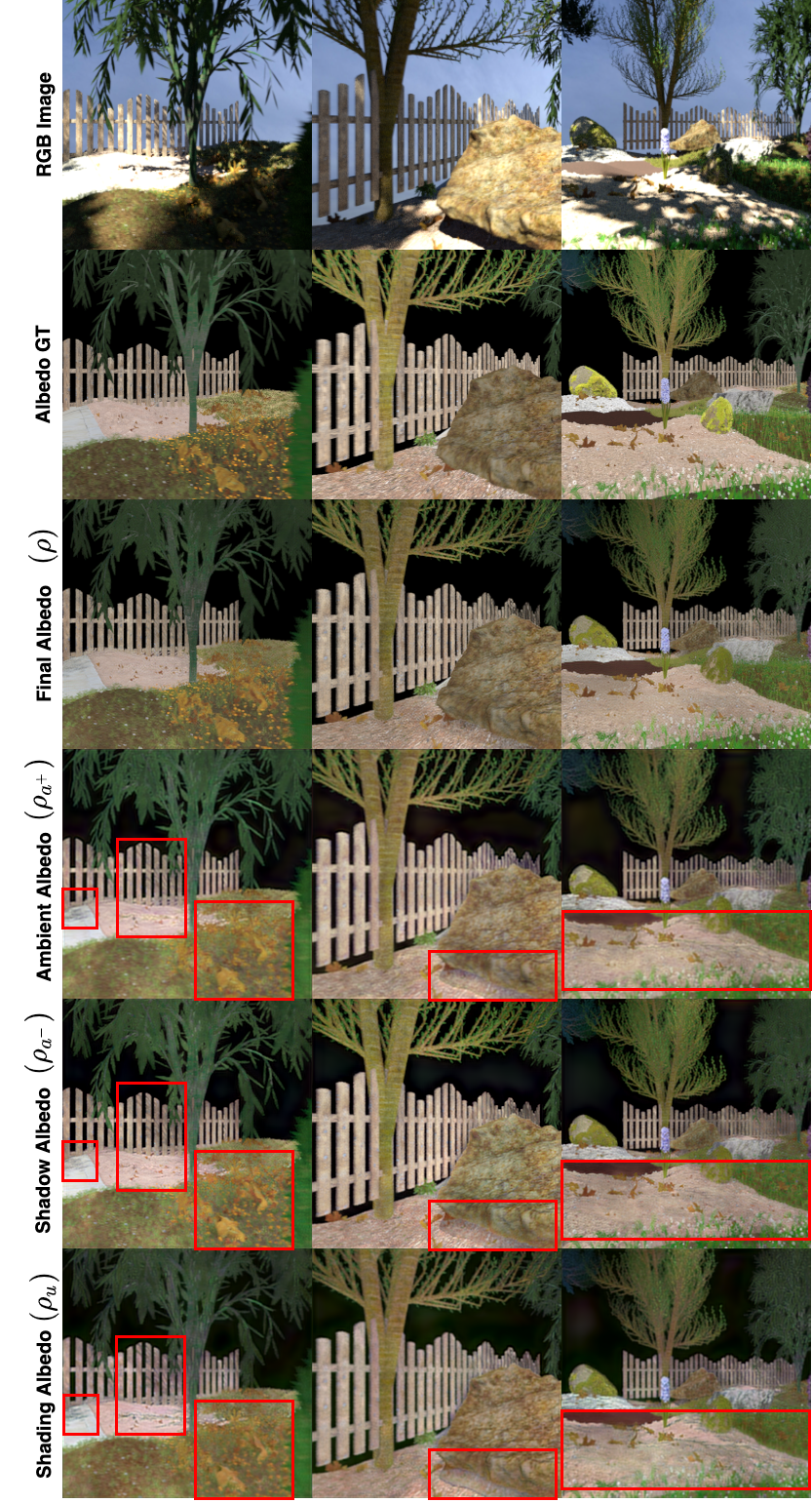}
\centering
\caption{Evaluation of the refinement module. Shading branch generated reflectance $\rho_{u}$ suffers the most and has the worst estimation quality. The refinement module further handles the shadow leakages and improves color augmentation for natural scenes.}
\label{fig:refinement}
\end{figure}

In addition, Figure~\ref{fig:refinement} provides a number of examples for NED. The qualitative results further demonstrate the benefits of the proposed refinement module. The 1st column demonstrates that the shading branch generated reflectance $\rho_{u}$ suffers the most and has the worst estimation quality. It is possible to see the strong shadow cast patterns on the ground below the tree and on the fences. Likewise, 2nd and 3rd columns show that the shading branch generated reflectance generations cannot properly handle photometric cues and they are contaminated with strong shadow cast patterns. There is no significant visual difference between $\rho_{a^{+}}$ and $\rho_{a^{-}}$ estimations. Nonetheless, they are better than $\rho_{u}$ predictions with less shadow leakages. On the other hand, the reflectance map generated by the refinement module is very close to the ground-truth images. The module further handles the shadow leakages, generates sharper estimations, improves color augmentation, and it can capture fine-grained thin objects such as the orange flowers in the 1st column. Visual evaluations for real world images without ground-truths are provided in Appendix~\ref{app:refinement_real}.

\subsection{Evaluation of the Fine-Grained Shadings}
\label{sec:photometric_ablation}
In this section, we evaluate the reconstruction qualities of the fine-grained shading estimations on the extended NED. The baselines are provided as comparisons. Quantitative results are provided in Table~\ref{tab:fine_grained_eval}. It can be observed that \emph{Baseline-a} conditioning the photometric predictions to a single unified shading decoder achieves better results than \emph{Baseline-b} with individual decoders. On the other hand, our model significantly achieves better reconstruction quality over the baselines for all components. Even though we do not predict direct shading component and use it as a self supervision signal, we also achieve better reconstruction quality on direct shading map generation. That also highlights the importance of our design choices. Among all the fine-grained shading components, the direct shading component appears to be the most challenging for all models.

\begin{table}[h]
   \centering
   \caption {Evaluation of the fine-grained shadings. We significantly achieve better reconstruction quality for all components on all metrics. $e^{+}_a$ is for ambient light, $e^{-}_a$ is for shadow casts, and $e_d$ is for direct shading.}
   \scalebox{0.9}{
   \begin{tabular}{|l||c|c|c|}\hline
      & SMSE ($e^{+}_a$) & SMSE ($e^{-}_a$) & SMSE ($e_d$) \\ \hline
     Baseline-a & 0.0155 & 0.0256 & 0.0545 \\ \hline
     Baseline-b & 0.0162 & 0.0293 & 0.0579 \\ \hline \hline
     ShadingNet (Ours) & \textbf{0.0103} & \textbf{0.0209} & \textbf{0.0459} \\ \hline
   \end{tabular}}
    \label{tab:fine_grained_eval}
 \end{table}

 \begin{figure}[h]
\includegraphics[width=1\linewidth]{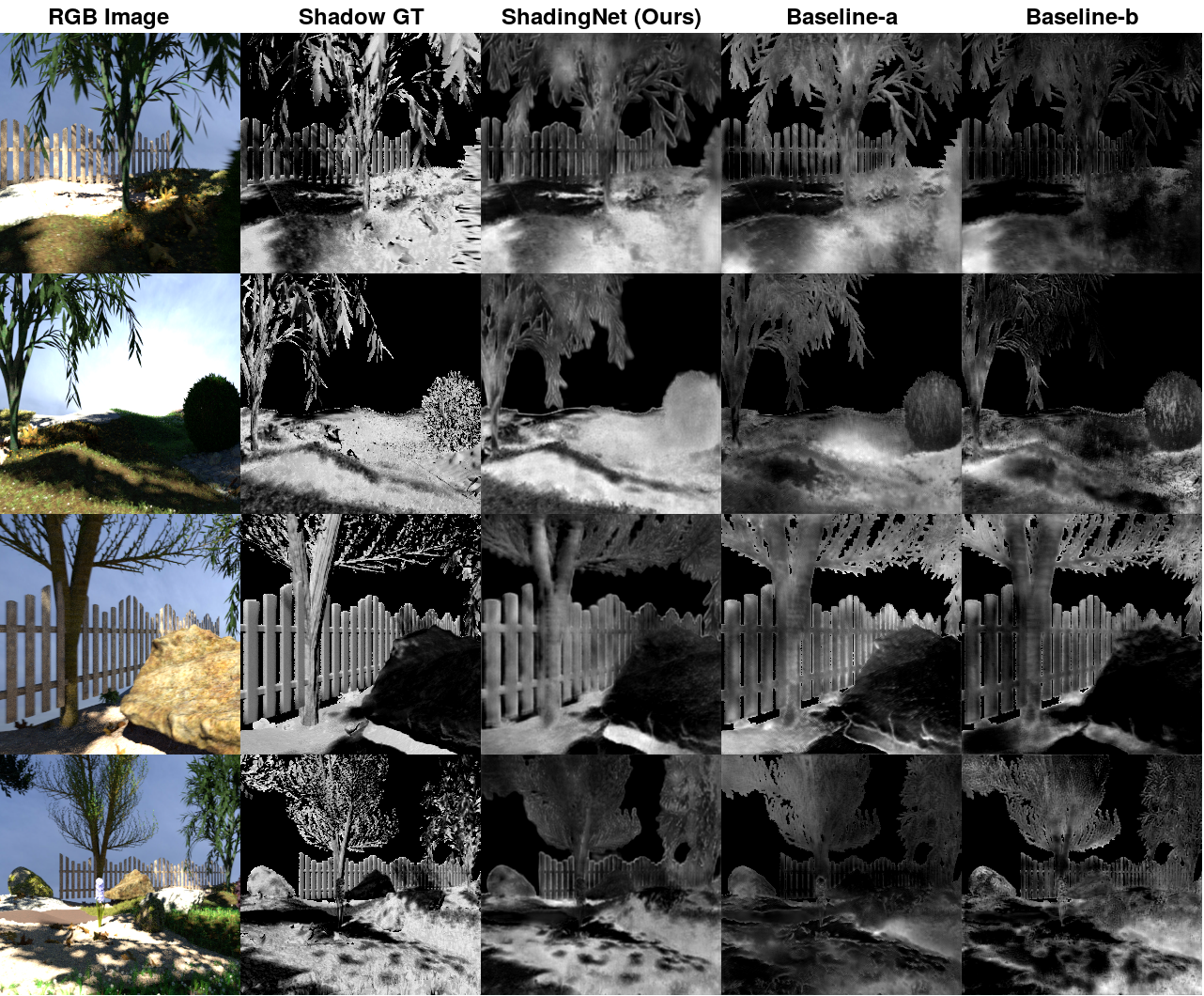}
\centering
\caption{Shadow estimations on NED. Our model is aware of diverse patterns and generates sharper maps.}
\label{fig:ned_shadow}
\end{figure}

 \begin{figure}[t]
\includegraphics[width=1\linewidth]{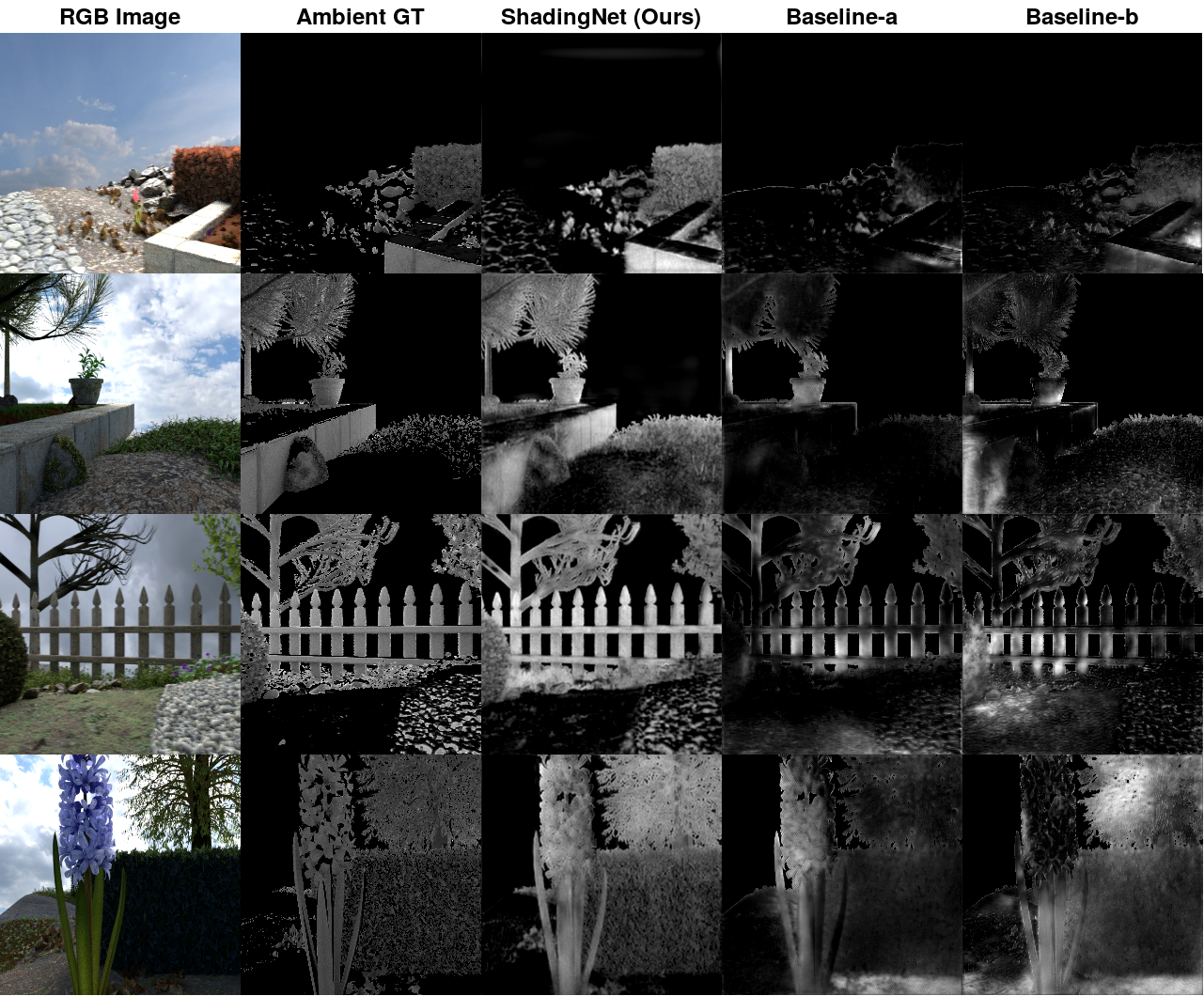}
\centering
\caption{Ambient light estimations on NED. Our model is aware of the regions that the direct light cannot reach.}
\label{fig:ned_ambient}
\end{figure}

 \begin{figure}[t]
\includegraphics[width=1\linewidth]{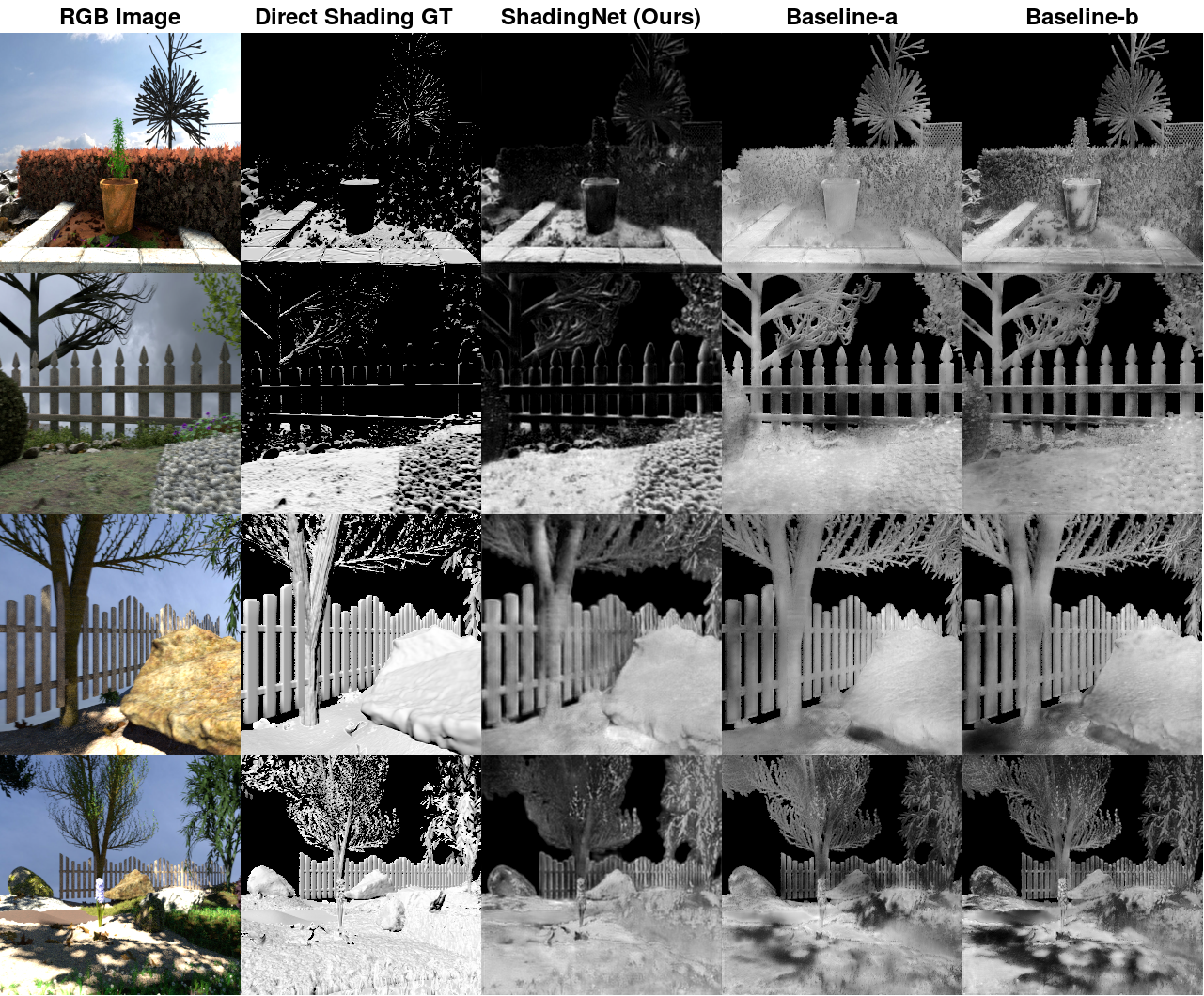}
\centering
\caption{Direct shading estimations on NED. Our model is better aware of the light properties, texture cues and the cast shadow features.}
\label{fig:ned_direct_shading}
\end{figure}

 \begin{figure*}[t]
\includegraphics[width=1\linewidth]{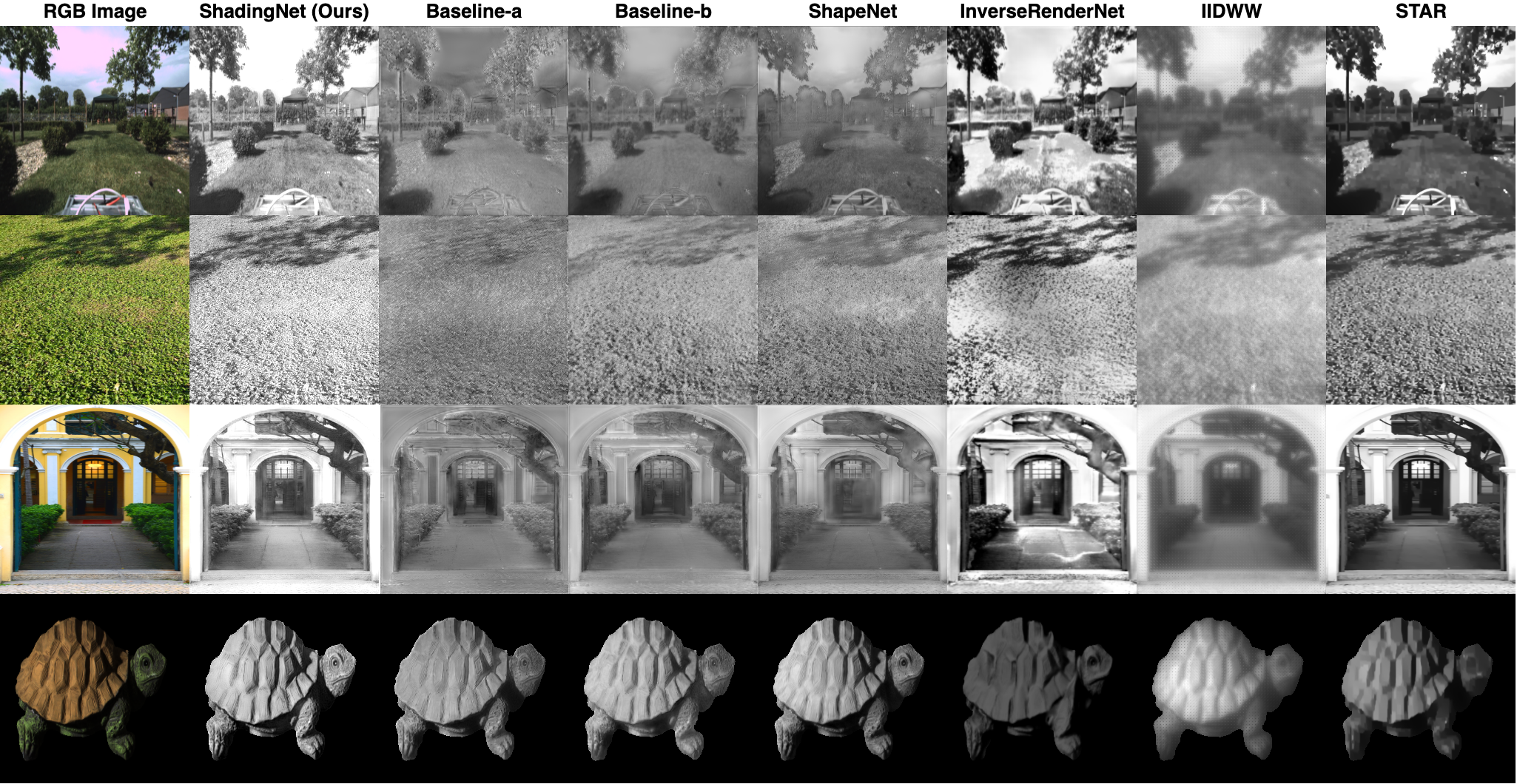}
\centering
\caption{Shading evaluations on real world objects. ShadingNet generates better shading maps that are sharper and it can properly capture the photometric cues. Images are best viewed on the electronic version.}
\label{fig:shading}
\end{figure*}

In addition, Figure~\ref{fig:ned_shadow} provides a number of examples for the shadow cast estimations. The baselines are able to handle certain cues. \emph{Baseline-a} estimations appear relatively better. Our model is aware of diverse shadow patterns and the estimations are closer to the ground-truth images and sharper than the baselines. Additionally, Figure~\ref{fig:ned_ambient} provides a number of examples for the ambient light estimations. Our model is aware of the indirect light cues and it can recognize the regions that the direct light cannot reach. The baselines tend to fail most of the cases, yet the 4th row demonstrates that they perform relatively reasonable on low light handling. Similar to the shadow cast evaluations, \emph{Baseline-a} estimations appear better than \emph{Baseline-b}. On the other hand, our ambient light estimations are significantly superior than the baselines. Finally, Figure~\ref{fig:ned_direct_shading} provides a number of examples for the direct shading estimations. The 1st row shows that \emph{Baseline-a} is aware of the texture cues of the pot, but cannot properly assess the light source position. Likewise, they fail to capture the light source position in the 2nd row and the rock is not visible anymore. The last two rows show that the estimations are polluted with (indirect) shadow cast features. On the other hand, our model is better aware of the light source position, texture cues and the cast shadow features. Our method still makes mistakes, such as the shadow patterns of the rock (3rd row), and the fence, rock on the left and ground (4th row) still contain shadow cues. This can be attributed to our self-supervision mechanism that the direct shading component is calculated by Equation~\eqref{eq:shading_formation} using the estimated shading components. Therefore, the individual errors might be accumulated. Furthermore, the quantitative results have shown that the direct shading component appears to be the most challenging component for all models. The reason might be that the geometry and lighting information are entangled in the representation and the ground-truth component is calculated by Equation~\eqref{eq:SF} using the surface normals and the light source properties, whereas the estimations are extracted from single $RGB$ images without any explicit supervision or regularization on surface normal features or light source properties. 

\subsection{Shading Estimations}
\label{sec:shading}
So far, we have focused on the reflectance predictions and fine-grained shading components. The experiments have shown that explicitly modelling the photometric cues further improves the reflectance estimation qualities. We have also provided quantitative evaluations for shading estimations when ground-truth labels are available. In this section, we provide a number of qualitative evaluations of the shading estimations on real world images in Figure~\ref{fig:shading}. The baselines generate rather blurry shading maps. The robot is not clearly visible in the 1st row and in the 2nd row \emph{Baseline-a} fails to capture the crisp shadows. There is no significant difference between the baselines and ShapeNet backbone. InverseRenderNet generates additional undesired shadow like artifacts. IIDWW estimations are oversmoothed and too blurry due to its smoothness priors. STAR achieves more decent estimations than other methods. Its shading estimations are remarkably better than the reflectance ones. On the other hand, ShadingNet generates better shading maps that are sharper and it can properly capture the photometric cues compared with the other learning-based methods. Compared with STAR, our generations are rather sharper in the first two rows, yet STAR estimations appear better in terms of contrast. 

\section{Conclusion}
\label{sec:conclusion}
Our aim was to improve the reflectance image estimation quality by explicitly modelling the photometric cues. To achieve that, the standard (Lambertian) image formation model was extended to incorporate the fine-grained shading components. The shading component is further factorized into different photometric effects such as shading caused by direct shading (object geometry) and indirect shading (shadows and ambient light) to generate better reflectance maps for natural scenes. An end-to-end supervised CNN model, \emph{ShadingNet}, were utilized to exploit the fine-grained model. The model was specifically designed to improve the reflectance estimations with special decoders, soft attention mechanisms and a novel refinement module. Since we are the first work to estimate fine-grained shading intrinsics, we extended two versions of a state-of-the-art intrinsic image decomposition model as baselines to provide a fair comparison. Along with the two baselines exploiting the fine-grained model, the performances of four supervised state-of-the-art deep learning models utilizing a unified shading map were evaluated. Furthermore, to train the models, a large-scale dataset of synthetic images of outdoor natural environments (NED) was extended generating fine-grained intrinsic images. All the models were trained on the extended NED. Then, three unsupervised learning based methods, including an outdoor inverse rendering and a model trained on image sequences also included for the evaluations. Finally, a structure and texture aware optimization based advanced Retinex was evaluated. The evaluations were provided on seven different datasets (NED, MPI Sintel, GTA V, IIW, MIT Intrinsic Images, 3DRMS and SRD) with comprehensively different setups without any fine-tuning or domain adaptation stage.

The evaluations prove that intrinsic image decomposition highly benefits from the proposed fine-grained shading model. Explicitly classifying the photometric effects significantly improves the reflectance estimations. For most of the cases, the baselines with fine-grained estimations further improve the backbone model predicting a unified shading component. On the other hand, our proposed ShadingNet constantly outperforms the baselines having fine-grained shadings, state-of-the-art supervised models predicting a unified shading, the outdoor inverse rendering method, the model trained on image sequences and an advanced Retinex method by a large margin. The qualitative comparisons demonstrated that ShadingNet properly handles diverse shadow patterns, low-light environments and strong shading effects. It is able to generate reflectance maps that are sharper, more natural and vivid with proper color augmentation and reproduction. In addition, it can generate sharper and richer shadow maps with various shadow patters, and it is aware of achromatic surfaces. Our model can also better differentiate brightness changes. It can detect the source of the bright pixels and can decently attribute them to reflectance or ambient light maps. Finally, our model emerges more stable. It presents an exceptional generalization performance with the ability to properly handle synthetic scenes, in-the-wild complex natural scenes and also object-level images.

\begin{acknowledgements}
This project was funded by the EU Horizon 2020 program No. 688007 (TrimBot2020).
\end{acknowledgements}

\section*{Conflict of interest}
The authors declare they have no conflict of interest.

\appendix
\section*{Appendix}
\section{Implementation Details of the Baselines}
\label{app:baseline_training_details}
The training of the baselines aligns with the implementation details of the ShapeNet backbone. All filter weights are initialized using a normal distribution. Adadelta optimizer with learning rate of $0.01$ is utilized \citep{Zeiler2012}. The learning rate is decayed until $1e-5$. The input images are normalized to the range of $[0, 1]$. Following the backbone, the models are trained until convergence using the scale invariant mean squared error. Similar to ShadingNet, one $\mathcal{L}_{c}$ is assigned to each intrinsic component, yielding 4 distinct loss functions (reflectance, direct shading, ambient light and shadows). Finally, all loss functions are added up without any weight tuning (all the weights are set to 1). Finally, following the backbone implementation, pixel errors are re-weighted with image gradients to generate more accurate and sharp edges. 

\section{Additional Qualitative Results on IIW}
\label{app:iiw_visuals}
\begin{figure}[t]
\includegraphics[width=1\linewidth]{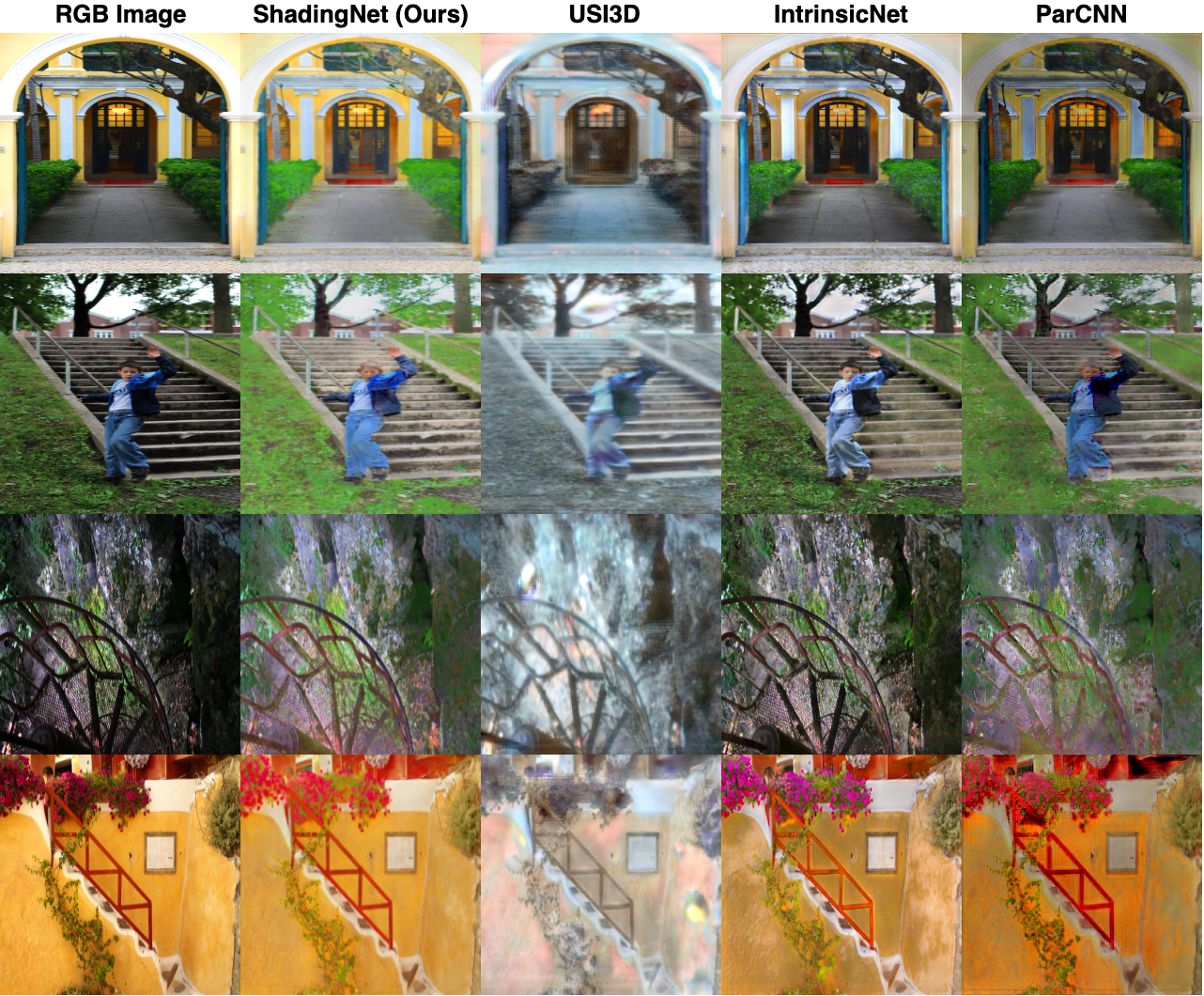}
\centering
\caption{Additional qualitative comparisons on IIW. Our method achieves superior results also on real world complex outdoor images.}
\label{fig:iiw_app}
\end{figure}

Additional qualitative comparisons for IIW are provided in Figure~\ref{fig:iiw_app}. IntrinsicNet generated reflectance maps are more vivid than ShapeNet variants and the stairs of the 2nd row displays a better shading handling performance than others. However, it generates color artefacts on the last row. ParCNN estimations are rather blurry and color artefacts are detectable. USI3D fails to capture proper colors. It generates rather dull reflectance images with artefacts. Moreover, it does not display decent shading handling performance.

\section{Additional Qualitative Results on SRD}
\label{app:srd_visuals}
\begin{figure}[t]
\includegraphics[width=1\linewidth]{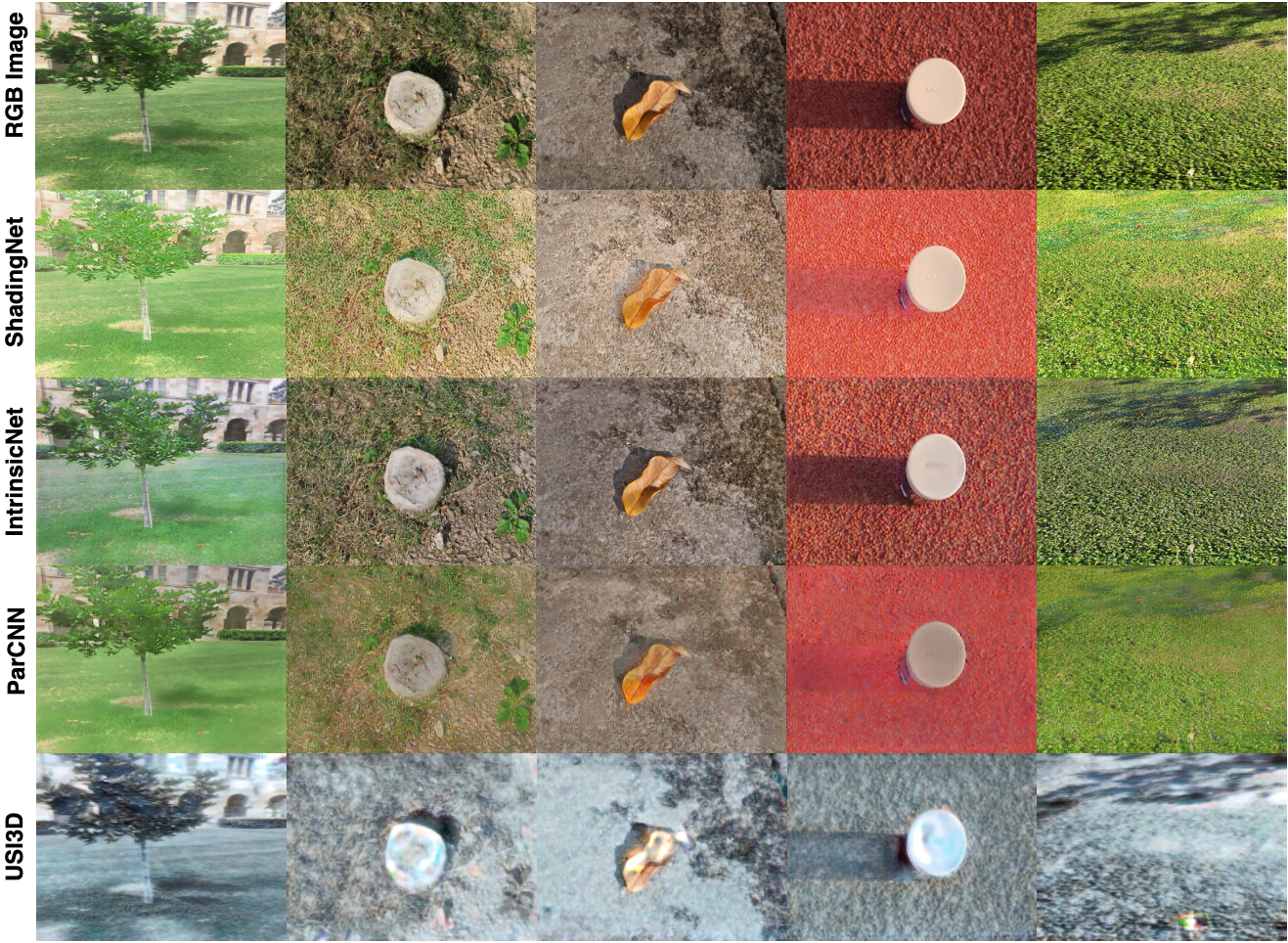}
\centering
\caption{Additional qualitative comparisons on USR. Our proposed ShadingNet generates significantly better reflectance maps that the colors appear more natural and vivid, and the structures are well-preserved. Other learning-based methods cannot properly handle shadow casts and some even generate additional artefacts.}
\label{fig:srd_app}
\end{figure}

Additional qualitative comparisons for SRD are provided in Figure~\ref{fig:srd_app}. IntrinsicNet estimations extremely resemble the input $RGB$ images, yet it can be observed that the direct light effects are slightly smoothed out. ParCNN generates (color-wise) relatively better reflectance images than other models. USI3D fails to capture proper colors. It generates rather dull reflectance images and all contaminated with shadow cues.

\section{Additional Qualitative Refinement Results on Real World Images}
\label{app:refinement_real}
\begin{figure}[h]
\includegraphics[width=1\linewidth]{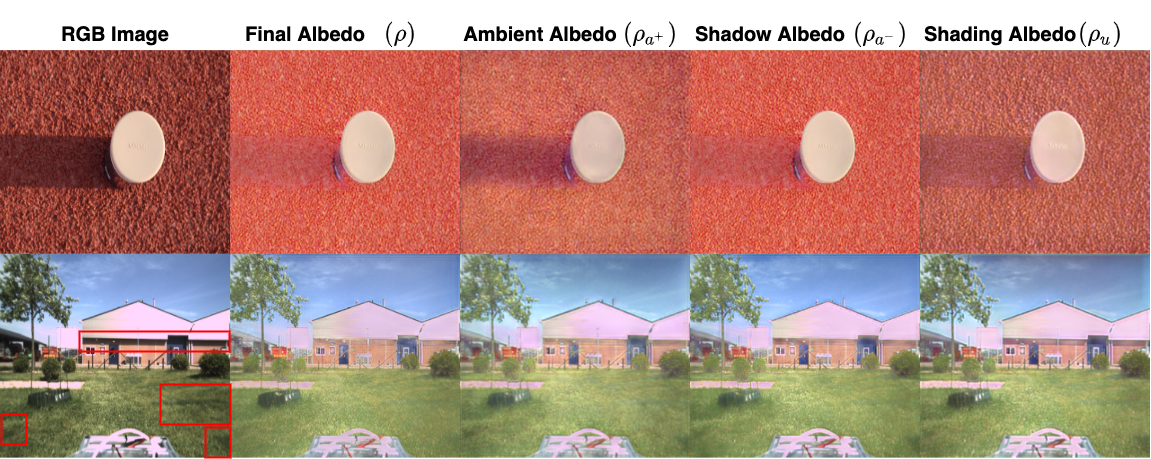}
\centering
\caption{Evaluation of the refinement module. Shading branch generated reflectance $\rho_{u}$ suffers the most and has the worst estimation quality. The refinement module further handles the shadow leakages, generates sharper estimations and improves color augmentation.}
\label{fig:refinement_real}
\end{figure}

Figure~\ref{fig:refinement_real} provides examples for SRD (1st row) and 3DRMS (2nd row) real world outdoor scenes to evaluate the refinement module. The 1st row demonstrates that $\rho_{u}$ and $\rho_{a^{+}}$ have the worst estimation quality. The reflectance map generated by the refinement module has minimum shadow leakage and the colors appear more vivid and natural. The 2nd row repeatedly shows that the shading branch generated reflectance $\rho_{u}$ suffers the most and has the worst estimation quality. The refinement module further handles the shadow leakages and generates relatively smoother estimations.

\begin{figure}[h]
\includegraphics[width=1\linewidth]{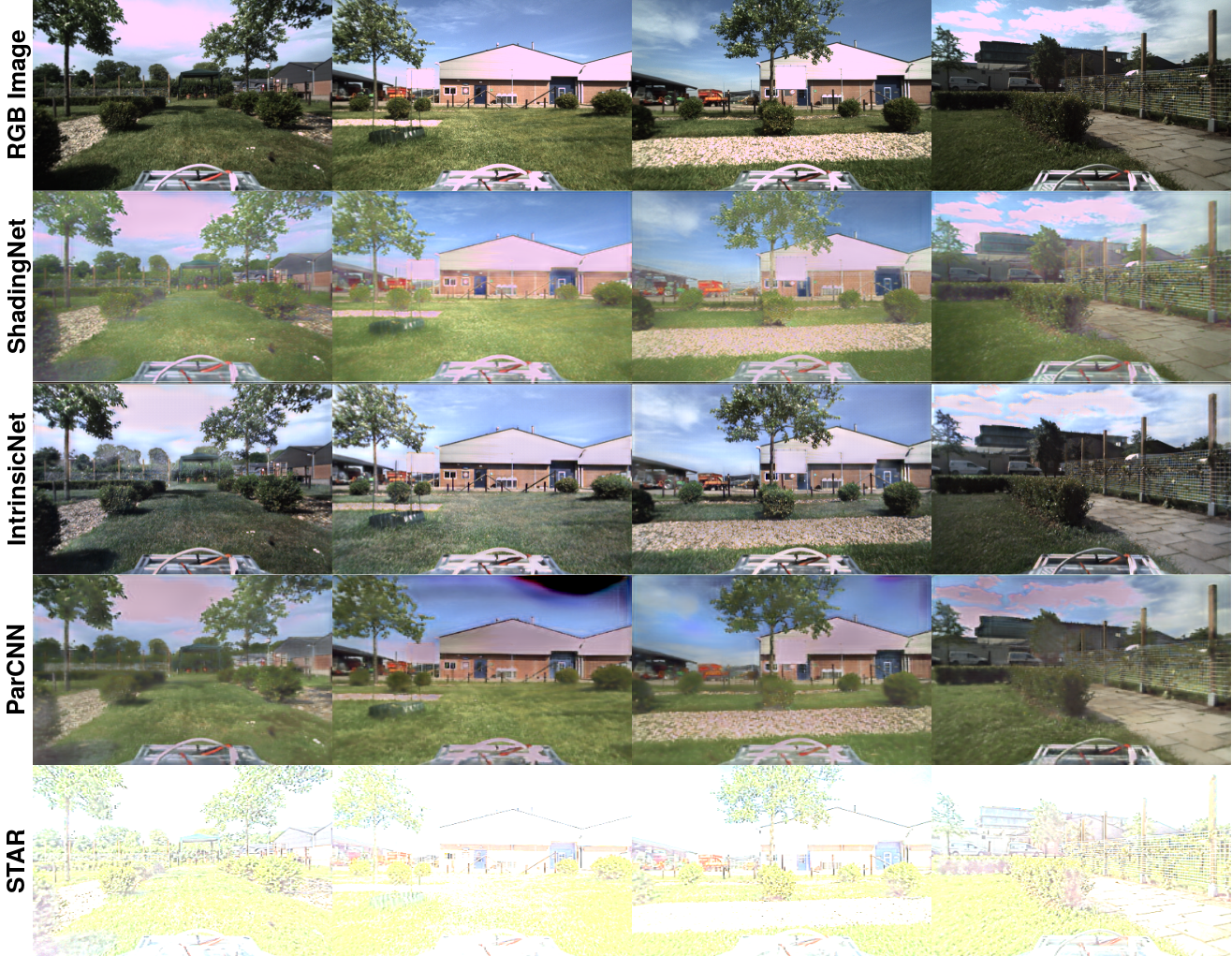}
\centering
\caption{Additional comparisons on 3DRMS. ShadingNet generates significantly better reflectance maps with more natural and vivid colors, and the structures are well-preserved. We perform better on handling various shadow patterns and also low light conditions.}
\label{fig:trimbot_app}
\end{figure}

 \begin{figure*}[t]
\includegraphics[width=1\linewidth]{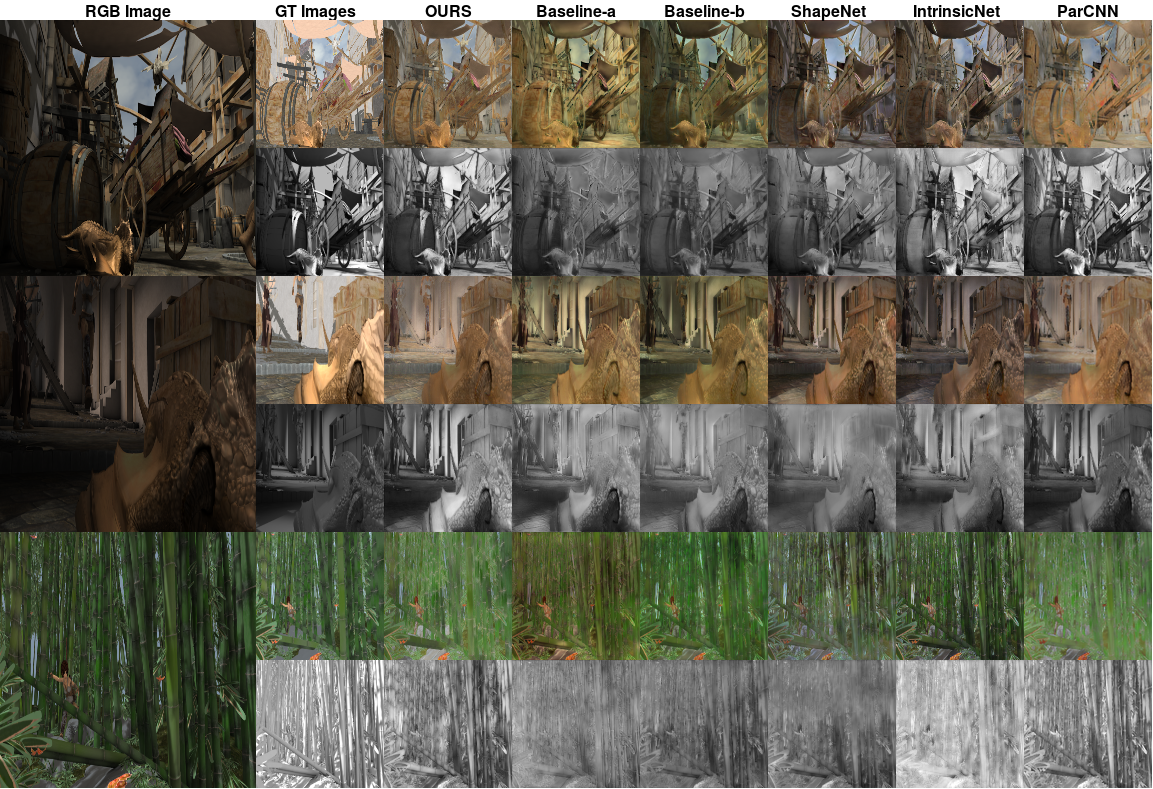}
\centering
\caption{Qualitative estimation results on MPI Sintel. Our proposed ShadingNet generated reflectance maps are closer to the ground-truth ones with minimal shadow artefacts. We also perform better on shadow and low-light handling. Similarly, our shading generations are sharper and closer to the ground-truth ones. A significant visual difference is not observed between the baselines and ShapeNet backbone. Best viewed on the electronic version.}
\label{fig:sintel}
\end{figure*}

\section{Additional Qualitative Results on 3DRMS}
\label{app:trimbot_visuals}
Additional qualitative comparisons for 3DRMS are provided in Figure~\ref{fig:trimbot_app}. IntrinsicNet model fails to generate proper colors and wrongly classifies shadows cues into reflectance predictions. It can smooth out the direct light effects, but clearly fails on handling indirect light cues. ParCNN generates reflectance images better than IntrinsicNet, but tends to fail on the sky regions (2nd and 3rd columns) generating purple/black artefacts and contaminating the reflectance generations. STAR model can handle a variety of the shadow casts, but by doing so it generates reflectance images that are way too bright that most of the structures and colors are not visible anymore. For example, in the 2nd column it appears that the irregular terrain filled with wood chips are falsely further extended to right with the white color artefacts and the color of the buildings are not recognizable.

\section{Qualitative Evaluations on MPI Sintel}
\label{app:sintel_visuals}
The qualitative comparisons for MPI Sintel are provided in Figure~\ref{fig:sintel}. It shows that the proposed ShadingNet generates reflectances that are closer to the ground-truth ones with minimal shadow artefacts. We also perform better on shadow and low-light handling. Similarly, our shading generations are sharper and closer to the ground-truth ones. A significant visual difference is not observed between the baselines and ShapeNet backbone. ParCNN appears color-wise better than other models, yet it does not perform as well as others on shadow handing.

\section{Qualitative Shading Evaluations on NED}
\label{app:ned_visuals}
\begin{figure*}[!h]
\includegraphics[width=1\linewidth,height=0.4\textheight]{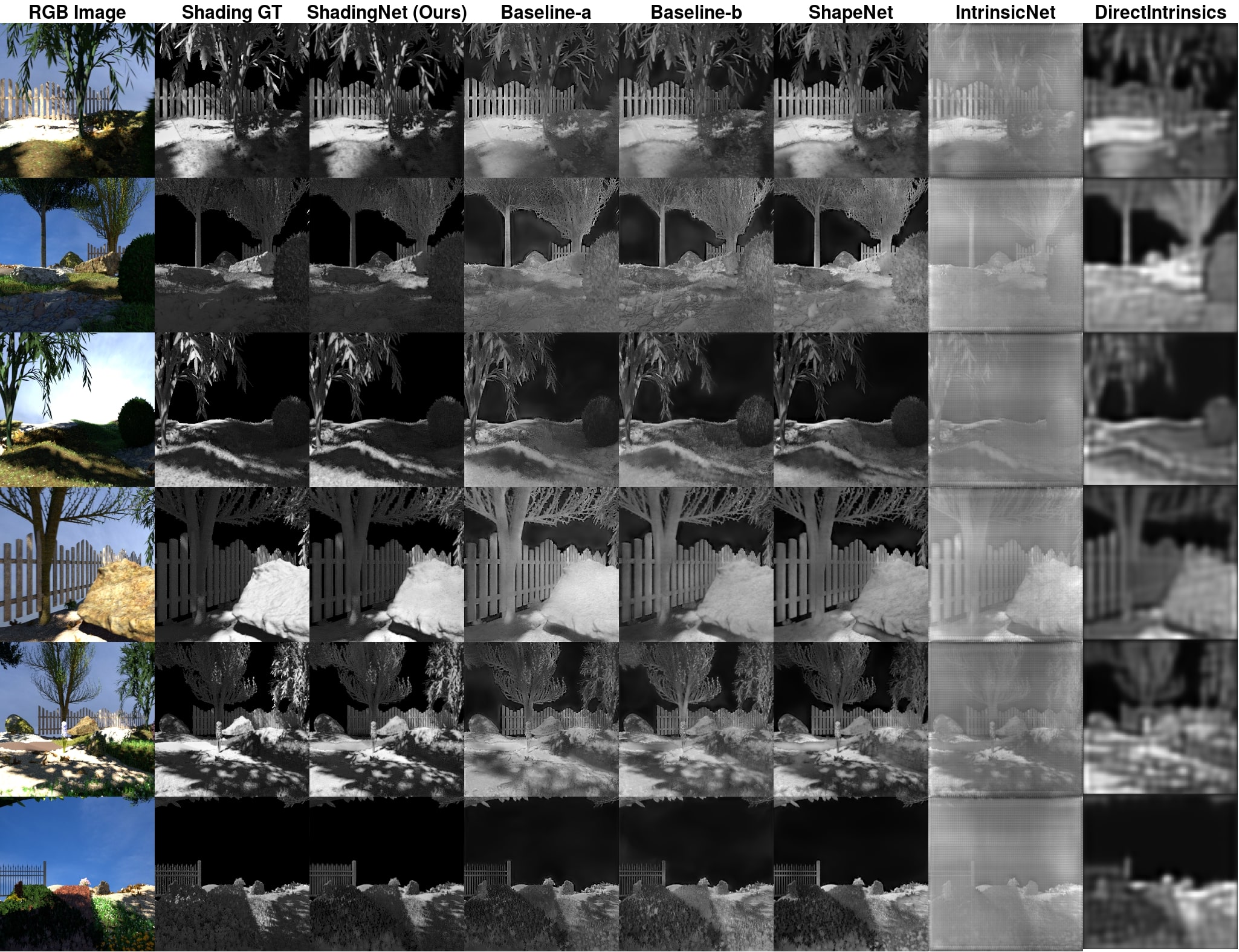}
\centering
\caption{Qualitative shading estimation results on NED's test set. ShadingNet produces sharper outputs that are very close to the ground-truth images. A significant visual difference is not observed among other models. IntrinsicNet estimations are problematic with checkerboard artefacts. DirectIntrinsics generations appear too blurry. }
\label{fig:ned_shading}
\end{figure*}

The qualitative comparisons for shading estimations on NED's test set are provided in Figure~\ref{fig:ned_shading}. It shows that the proposed ShadingNet produces sharper shading maps that are very close to the ground-truth images. A significant visual difference is not observed among other models. IntrinsicNet shading estimations are better than its reflectance estimations, but again problematic with checkerboard artefacts. Likewise, DirectIntrinsics generations are too blurry.

\begin{figure*}[!h]
\includegraphics[width=1\linewidth,height=0.35\textheight]{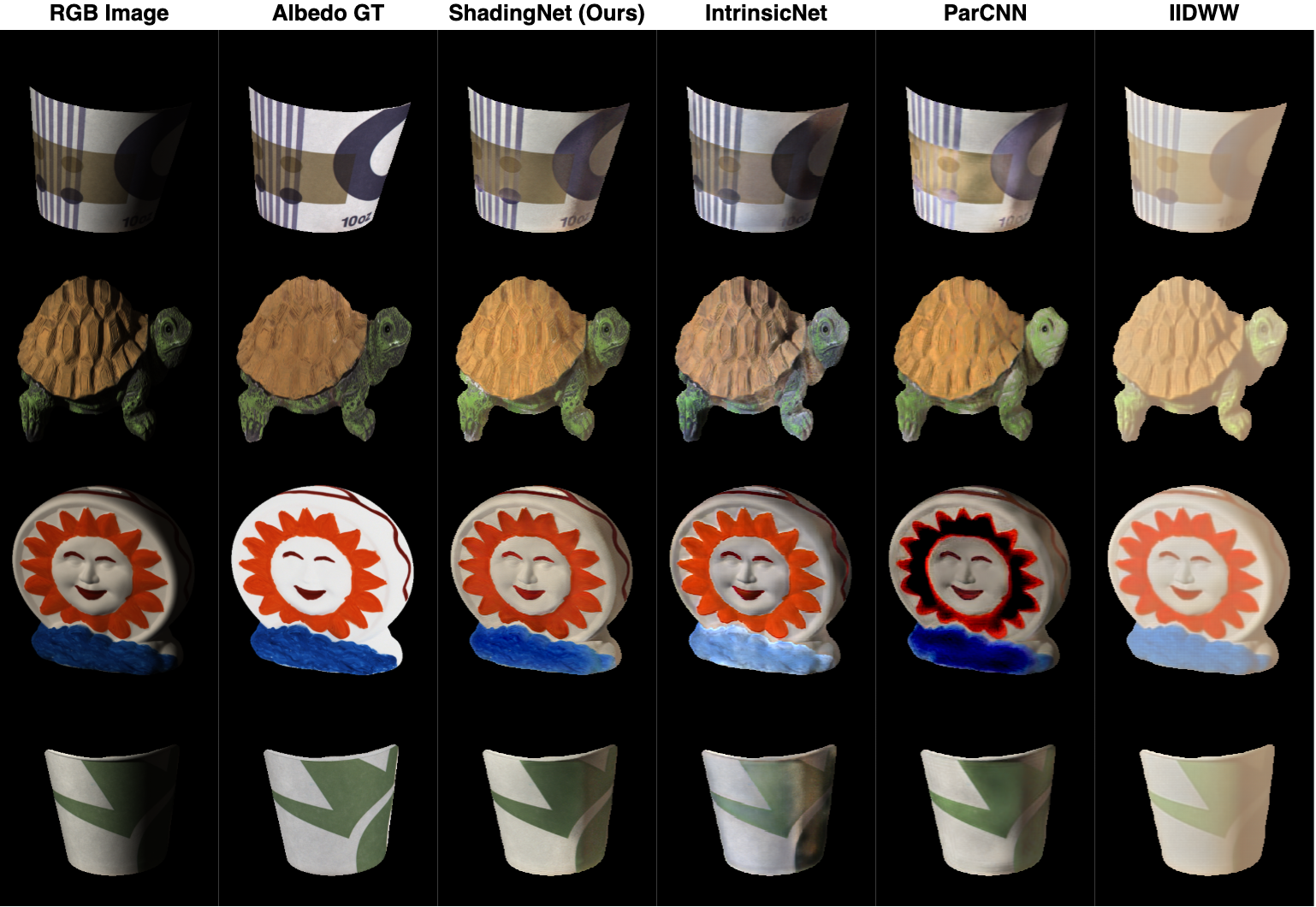}
\centering
\caption{Additional comparisons on MIT. We generate sharper estimations that are closer to the ground-truths with better color reproduction and shading handling.}
\label{fig:mit_app}
\end{figure*}

\section{Additional Qualitative Results on MIT}
\label{app:mit_visuals}
Additional qualitative comparisons for MIT are provided in Figure~\ref{fig:mit_app}. IntrinsicNet estimations have less artefacts than others, but the model cannot properly handle the strong shadings. ParCNN generates additional brightness artefacts, wrongly eliminates the black dots on the head of the turtle and confuses them with shading cues, and completely fails on the sun image. IIDWW generated colors are dull and the images contain yellowish color cast. Further, the model cannot handle strong shadings due to geometry such as the right-hand parts of the objects where the curvature is the strongest and the farthest from the light source.

\bibliographystyle{spbasic}      % basic style, author-year citations
\bibliography{refs}   % name your BibTeX data base

\end{document}